\documentclass{article} % For LaTeX2e
\usepackage[preprint]{colm2026_conference}
\usepackage{microtype}
\usepackage{hyperref}
\usepackage{url}
\usepackage{booktabs}
\usepackage{graphicx}
\usepackage{xcolor}
\usepackage{multirow}
\usepackage{colortbl}
\usepackage{lineno}
\usepackage{subcaption}
\usepackage{amsmath,amssymb}
\usepackage{pifont}
\usepackage{algorithm}
\usepackage{algorithmic}
\usepackage{wrapfig}
\definecolor{darkblue}{rgb}{0, 0, 0.5}
\usepackage{tikz}
\usetikzlibrary{decorations.pathreplacing}

\usepackage[T1]{fontenc}
\usepackage{listings}
\usepackage{xcolor}
\usepackage{tikz}
\usepackage{booktabs}
\usepackage{enumitem}
\lstdefinestyle{workflow}{
  language=Python,
  basicstyle=\small\ttfamily,
  keywordstyle=\bfseries\color{blue!70!black},
  stringstyle=\color{red!60!black},
  commentstyle=\itshape\color{green!50!black},
  breaklines=true,
  frame=single,
  framesep=6pt,
  xleftmargin=8pt,
  xrightmargin=8pt,
  aboveskip=10pt,
  belowskip=10pt,
  numbers=left,
  numberstyle=\tiny\color{gray},
  numbersep=6pt,
  showstringspaces=false,
  tabsize=4,
  morekeywords={await, async, self},
}

\hypersetup{colorlinks=true, citecolor=darkblue, linkcolor=darkblue, urlcolor=darkblue}

\newcommand{\method}{SWIFT}
\newcommand{\methodfull}{Synthesizing Workflows via Few-shot Transfer}

\title{Why Search When You Can Transfer? Amortized Agentic Workflow Design from Structural Priors}

\author{
\begin{tabular}{l}
\textbf{Shiyi Du}$^{1}$, \textbf{Jiayuan Liu}$^{1,2}$, \textbf{Weihua Du}$^{1}$, \textbf{Yue Huang}$^{3}$, \textbf{Jiayi Li}$^{1}$, \textbf{Yingtao Luo}$^{1}$,\\
\textbf{Xiangliang Zhang}$^{3}$, \textbf{Vincent Conitzer}$^{1,2}$, \textbf{Carl Kingsford}$^{1}$\thanks{Corresponding Author.}\\
\end{tabular}
\\
\begin{tabular}{l}
$^{1}$ Carnegie Mellon University \\
$^{2}$ Foundations of Cooperative AI Lab (FOCAL) \\
$^{3}$ University of Notre Dame \\
\end{tabular}
\\
\\
\begin{tabular}{l}
\small
\texttt{\{shiyid,\hspace{1pt}jiayuan4,\hspace{1pt}weihuad\}@andrew.cmu.edu},\ \ 
\texttt{yhuang37@nd.edu},
\texttt{\{lijiayi,\hspace{1pt}yingtaol\}@andrew.cmu.edu},\\
\small
\texttt{xzhang33@nd.edu},\ \
\texttt{conitzer@cs.cmu.edu},\ \ 
\texttt{carlk@cs.cmu.edu}\ \ 
\end{tabular}
}

\begin{document}

\ifcolmsubmission
\linenumbers
\fi

\maketitle

\begin{abstract}

Automated agentic workflow design currently relies on per-task iterative search, which is computationally prohibitive and fails to reuse structural knowledge across tasks. We observe that optimized workflows converge to a small family of domain-specific topologies, suggesting that this combinatorial search is largely redundant. Building on this insight, we propose \method{} (\methodfull{}), a framework that amortizes workflow design into reusable structural priors. \method{} first distills compositional heuristics and output-interface contracts from contrastive analysis of prior search trajectories across source tasks. At inference time, it conditions a single LLM generation pass on these priors together with cross-task workflow demonstrations to synthesize a complete, executable workflow for an unseen target task, bypassing iterative search entirely. On five benchmarks, \method{} outperforms the state-of-the-art search-based method while reducing marginal per-task optimization cost by three orders of magnitude. It further generalizes to four additional unseen benchmarks and transfers successfully from GPT-4o-mini to three additional foundation models (Grok, Qwen, Gemma). Controlled ablations reveal that workflow demonstrations primarily transfer topological structure rather than surface semantics: replacing all operator names with random strings still retains over 93\% of the full system's average performance.
\end{abstract}

% ===================== NEW INTRO (red) =====================
\section{Introduction}
\vspace{-5pt}
% {\color{red}

LLM agent workflows, which are structured programs that orchestrate sequences of LLM calls, tool uses, and control flow, have become central to solving complex reasoning and generation tasks~\citep{yao2022react, wei2022chain}. Manually designing such workflows is labor-intensive, prompting recent methods such as AFlow~\citep{zhang2025aflow} and ADAS~\citep{hu2024automated} to automate the process via iterative search (e.g., MCTS). While effective, these approaches share a fundamental limitation: they treat every new task as an independent optimization problem, redundantly exploring a vast discrete program space from scratch with no transfer of structural knowledge across tasks. This per-task redundancy is not only computationally prohibitive but may be actively counterproductive, as unconstrained search risks overfitting to small validation sets~\citep{el2025inefficiencies, wang2025efficient}.

A natural question arises: \emph{Is iterative search necessary for workflow design?} We observe that optimized workflows converge to a small family of domain-specific topologies: math tasks consistently favor multi-path
sampling with verification, while code tasks converge to execution-feedback retry loops. This suggests that the effective
search space is far more structured than the combinatorial paradigm assumes. If such structural regularities transfer across tasks, then the cost of workflow discovery can be amortized rather than paid independently for each new task.

\begin{figure*}[t]
    \centering
    \includegraphics[width=0.92\textwidth]{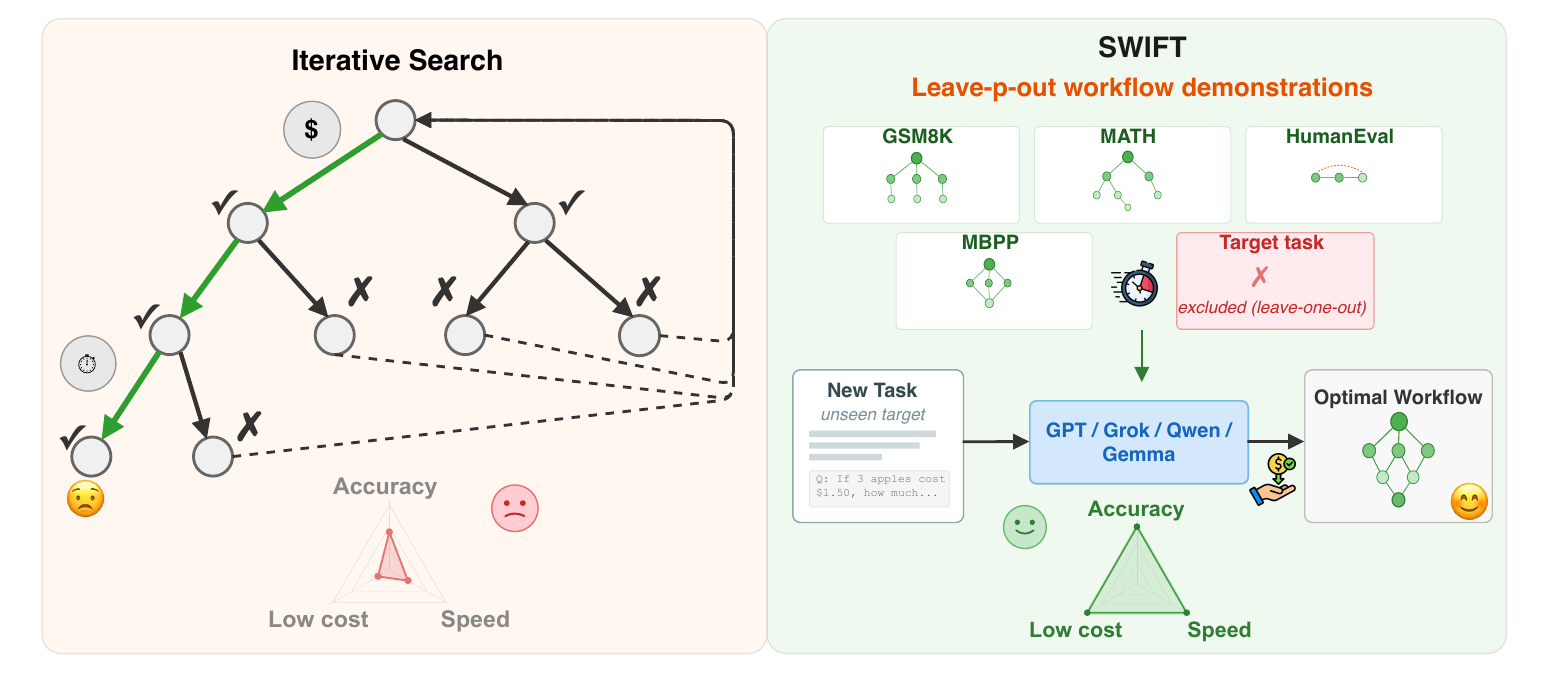}
 
\caption{Iterative search (left) vs.\ \method{} (right). }
    \label{fig:teaser}
    \vspace{-15pt}
\end{figure*}

We instantiate this idea in \method{}  (\methodfull{}; Figure~\ref{fig:teaser}), a framework that amortizes workflow design into reusable structural priors. \method{} operates in two phases. \textbf{Offline}, it performs \emph{contrastive trajectory distillation}: by analyzing both high- and low-performing workflows from prior search histories, it extracts compositional heuristics---actionable, operator-level routing rules that bound the generation space. It further distills \emph{output contracts} (strict formatting constraints that ensure the output of one node can be seamlessly parsed by the next) from intermediate-scoring workflows that failed due to formatting mismatches, capturing interface specifications that ensure executability. \textbf{Online}, given only a brief description of a new target task, \method{} synthesizes a complete, executable workflow for the target task in a single generation pass, conditioned on workflow examples from other tasks and a small set of automatically learned design rules. It involves no iterative search and no task-specific optimization.

To rigorously prevent information leakage, we adopt a strict leave-one-out (LOO) protocol: when generating a workflow for any target task, all data from that task is entirely masked.

Our main contributions are as follows:
\vspace{-5pt}
\begin{itemize}[leftmargin=*,itemsep=1pt,parsep=0pt,topsep=0pt]
    \item \textbf{Amortized workflow synthesis.} We reframe automated workflow design from per-task iterative search to amortized synthesis. \method{} distills reusable structural priors offline and synthesizes competitive workflows for unseen tasks in a single generation, reducing optimization cost by three orders of magnitude compared to the state-of-the-art search-based method.

    \item \textbf{Mechanistic understanding of cross-task transfer.} We show that contract-guided synthesis is essential for bridging the structural-functional gap, and that workflow demonstrations primarily transfer topological routing patterns rather than human-readable operator names (Table~\ref{tab:component_ablation}).

    \item \textbf{Strong empirical performance with broad generalization.} \method{} outperforms AFlow across all five main benchmarks (Table~\ref{tab:main_results}). It further generalizes to four entirely unseen benchmarks (Table~\ref{tab:ood_results}) and transfers successfully to three alternative foundation models---Grok, Qwen, and Gemma (Table~\ref{tab:generality}).

\end{itemize}

\section{Related Work}
\vspace{-5pt}

\paragraph{Structured reasoning and multi-agent workflows.}
Single-agent prompting strategies, chain-of-thought~\citep{wei2022chain}, complexity-based prompting~\citep{fu2022complexity}, self-consistency~\citep{wang2023selfconsistency}, tree-of-thought~\citep{yao2023tree}, and graph-of-thought~\citep{besta2024graph} all improve LLM performance through fixed reasoning patterns. Self-Refine~\citep{madaan2023self}, Reflexion~\citep{shinn2023reflexion}, and LATS~\citep{zhou2023language} further add iterative self-feedback or tree search at test time, but all operate per-instance.
Hand-crafted multi-agent systems explore diverse collaboration patterns like inter-agent debate~\citep{du2024improving}, output ensembling via ranking and fusion~\citep{jiang2023llm}, intra-model persona simulation~\citep{wang2024unleashing}, dynamic team selection~\citep{liu2023dynamic}, collaborative role adjustment~\citep{chen2023agentverse}, and graph-structured agent networks~\citep{qian2024scaling}, but all rely on manually specified topologies.
Recent studies corroborate the primacy of topology: \citet{zhou2025multi} find that communication topology matters more than prompt optimization, and G-Designer~\citep{zhang2024g} learns task-specific topologies via GNNs. \method{} shares this insight but \emph{synthesizes} topology via demonstration-conditioned generation rather than manual design or training a separate model.

\paragraph{Automated workflow optimization.}

Recent work automates agent design through search and optimization.
A large line of work focuses on \textbf{structure and workflow search}, 
including AutoAgents~\citep{chen2023autoagents}, GPTSwarm~\citep{zhuge2024gptswarm}, and AgentSquare~\citep{shang2024agentsquare}, 
which generate agent roles, computational graphs, or modular compositions via dynamic planning, graph optimization, or evolutionary search.
Related approaches such as AFlow~\citep{zhang2025aflow} and MaAS~\citep{zhang2025multi} further explore graph-based or probabilistic workflow spaces, 
though often requiring substantial per-task search cost.
Another line of work focuses on \textbf{prompt and program optimization}, 
including DSPy~\citep{khattab2024dspy}, OPRO~\citep{yang2023large}, and TextGrad~\citep{yuksekgonul2024textgrad}, 
which optimize prompts or pipelines via compilation, black-box optimization, or gradient-like feedback.
Finally, some approaches aim at \textbf{dynamic or learned composition}, 
such as DAAO~\citep{su2025difficulty} and MAS-GPT~\citep{ye2025mas}, 
which generate query-specific agent structures, often relying on fine-tuning or learned policies.
Despite these advances, prior work typically performs optimization independently for each task.
In contrast, \method{} amortizes structural knowledge across tasks, so that each new task benefits from prior optimization effort at near-zero marginal cost.

\paragraph{Demonstration-conditioned generation and cross-task transfer.}
LLMs can adapt to new tasks from few-shot demonstrations without parameter updates~\citep{brown2020language}. Recent work studies what makes demonstrations effective~\citep{min-etal-2022-rethinking, liu2022makes} and how to select them~\citep{rubin2022learning}. CrossICL~\citep{gao2025crossicl} transfers demonstrations across NLP tasks via unsupervised alignment, and \citet{wang2024context} synthesize target task demonstrations from similar source tasks. Agent Workflow Memory~\citep{wang2024agent} induces reusable workflow routines from past executions and retrieves them for new tasks. These methods transfer at the \emph{instance level}, selecting or adapting input-output examples for a given task. \method{} operates at the \emph{program level}: it uses complete executable workflows as demonstrations to synthesize a novel workflow in a single generation pass, without retrieving or adapting cached routines.

\paragraph{Amortized optimization and program synthesis.}
The idea of replacing per-instance optimization with a learned amortized mapping has deep roots in machine learning: amortized variational inference~\citep{kingma2013auto, rezende2014stochastic} replaces per-datapoint ELBO optimization with a shared encoder, and neural program synthesis~\citep{devlin2017robustfill, balog2016deepcoder} learns to predict programs from input-output specifications rather than searching over the program space. \method{} extends this amortization principle to the domain of agentic workflow design: rather than searching over graph topologies per task, it learns a direct mapping from task descriptions to executable workflows, conditioned on structural priors distilled from prior optimization trajectories.

\section{Methodology}
\vspace{-5pt}
\label{sec:method}
Automated LLM workflow generation currently relies on iterative search, which incurs prohibitive computational costs and fails to transfer knowledge across tasks. We propose \textbf{\method{}}, a framework that amortizes the combinatorial search cost into reusable structural priors. By conditioning workflow generation on cross-task demonstrations and distilled heuristics, \method{} enables few-shot workflow synthesis via a single generation pass. \textbf{The overall architecture of \method{} is illustrated in Figure~\ref{fig:architecture}.}

\begin{figure*}[hbpt]
\vspace{-5pt}
\centering
\includegraphics[width=0.95\textwidth]
{figures/swift_overview_updated5.pdf}
\caption{Overview of \method{}. \textbf{Offline}: contrastive trajectory distillation extracts compositional heuristics~($\mathcal{H}$) and output contracts~($\mathcal{C}$) from source-task search histories. \textbf{Online}: a single LLM call synthesizes an executable workflow for an unseen task, conditioned on operator definitions, distilled priors, and leave-one-out demonstrations. The workflow is directly executed on test instances.}

\label{fig:architecture}
\vspace{-15pt}
\end{figure*}

\vspace{-3pt}
\subsection{Problem Formulation: Workflow Synthesis as Graph Generation}
\vspace{-3pt}
\label{subsec:problem_formulation}

We formalize a workflow $w$ as an executable program that, given a task input $x$, orchestrates a sequence of operator calls to produce an output $y=w(x)$. Architecturally, $w$ is represented as an operator compositional graph $G = (\mathcal{V}, \mathcal{E})$, realized as executable program following~\cite{zhang2025aflow}. Each node $v_i \in \mathcal{V}$ is an instantiation of a predefined operator from a typed library $\mathcal{O}=\{o_1, \dots, o_K\}$, where each operator encapsulates a specific capability such as LLM reasoning, code execution, or ensemble voting~\citep{zhang2025aflow}. Crucially, each instantiated operator is parameterized by a natural language instruction $\theta_i$, and may encapsulate internal control flow such as conditional branching and retry loops. The edges $\mathcal{E}$ define the topological data flow and control logic between these operators.

Given a target task $\mathcal{T}_{target}$ and an evaluation dataset $\mathcal{D}$, the standard optimization objective seeks to find the optimal workflow $w^*$ that maximizes the expected task accuracy:
$$w^* = \arg\max_{w \in \mathcal{W}} \mathbb{E}_{x \sim \mathcal{D}} [\text{Acc}(w(x), \mathcal{T}_{target})]$$
Because the search space $\mathcal{W}$ is discrete, non-differentiable, and combinatorially vast, solving this from scratch per task via iterative algorithms (e.g., MCTS) requires $O(N \times C)$ cost, where $N$ is the number of search iterations and $C$ is the cost per LLM evaluation. Instead, \method{} learns an amortized mapping $f_\phi: \mathcal{T} \rightarrow \mathcal{W}$, where $\phi$ denotes the frozen parameters of a pretrained LLM that serves as the meta-generator, allowing it to deduce the optimal solver architecture in $O(1)$ inference steps.

\vspace{-3pt}
\subsection{Contract-Guided Synthesis to Bridge the Structural-Functional Gap}
\vspace{-3pt}
\label{subsec:contract_guided}
A failure mode inherent to meta-level demonstration-conditioned workflow synthesis is the \textit{structural-functional gap}. During the generation of structured programs, a synthesized workflow $w$ may be logically and topologically sound in its reasoning process (Structural) but violate the strict downstream evaluation interfaces (Functional). For example, a mathematically sound workflow might compute the correct numerical answer but return it embedded within conversational text, causing an exact-match evaluation parser to register a failure.

In a standalone LLM call, output formatting is trivially resolved by appending a format instruction. In a multi-node workflow, however, formatting errors cascade across node interfaces: an intermediate node may output semantically correct but syntactically incompatible text (e.g., ``The answer is 42'' instead of ``42''), which downstream nodes, ensemble voters, extraction operators, or evaluation parsers, cannot parse. Each node individually follows its local instruction, yet the inter-node handoff fails. Moreover, during amortized synthesis, the generator must produce a complete, novel workflow in a single generation pass without the opportunity for per-node prompt tuning. Without explicit interface constraints, the synthesis LLM compensates for this uncertainty by generating overly complex topologies: adding redundant ensemble voting, multi-candidate generation, and error recovery branches as hedging mechanisms. Our ablation confirms this effect empirically: removing contracts causes the synthesizer to produce workflows with substantially more nodes and higher per-sample cost, yet lower accuracy (Table~\ref{tab:component_ablation}; see Appendix~\ref{app:contract_topology} for a detailed topology comparison).

We resolve this phenomenon via \textbf{contract-guided synthesis}. We isolate execution logs from intermediate-scoring workflows in the search history, workflows that achieved partial success but failed final extraction. From these specific failure patterns, we distill a set of \textbf{Output Contracts ($\mathcal{C}$)}. These contracts act as formal interface specifications injected directly into the synthesis process, explicitly constraining the terminal nodes of the synthesized workflow to adhere to rigid evaluation formatting without relying on fragile, post-hoc string matching.

\vspace{-3pt}
\subsection{Demonstration-Conditioned Workflow Synthesis}
\vspace{-3pt}
\label{subsec:zero_shot_synthesis}

Equipped with the distilled priors $\mathcal{H}$ and $\mathcal{C}$, the synthesis of new workflows is reduced to a demonstration-conditioned generation problem. For the unseen target task $\mathcal{T}_{target}$, we construct a structured meta-prompt $P$ that encapsulates the available action space (the operator library $\mathcal{O}$), the distilled structural and functional constraints ($\mathcal{H} \cup \mathcal{C}$), and a retrieved set of high-performing structural demonstrations $\mathcal{D}_{demo} \subset \{ w_1^*, \dots, w_N^* \}$.

Under our LOO protocol, $\mathcal{D}_{demo}$ is strictly disjoint from $\mathcal{T}_{target}$. These cross-task demonstrations condition the generator via two decoupled channels: (1)~\textbf{Syntax Specification}, providing a strict schema definition for class instantiation and Python execution; and (2)~\textbf{Strategy Transfer}, transferring higher-order cognitive architectures across task boundaries, independent of the semantic content of the target task.

By conditioning on $P$, the meta-generator executes $w_{target} \leftarrow f_\phi(P)$, implicitly performing an architectural search in a single generation pass and outputting the executable workflow. The complete procedure is formally described in Algorithm~\ref{alg:amortized_synthesis} (Appendix~\ref{app:algorithm}).

\section{Experiments}
\vspace{-5pt}
\label{sec:experiments}

To evaluate the efficacy of \method{} as an amortized workflow synthesis framework, we design our experiments to move beyond standard performance benchmarking to empirically answer the following research questions:
\vspace{-5pt}
\begin{itemize}[leftmargin=*,itemsep=1pt,parsep=0pt,topsep=0pt]
    \item \textbf{RQ1 (Efficacy \& Task Transfer):} Can single-pass, demonstration-conditioned synthesis match or exceed computationally expensive iterative search (e.g., MCTS)? Do the meta-learned structural priors generalize to entirely out-of-distribution (OOD) task families?
    \item \textbf{RQ2 (The Nature of the Search Space):} Does the $O(1)$ synthesis cost of \method{} suggest that the optimization landscape of agent workflows is structurally smooth across tasks, rendering combinatorial search redundant?
    \item \textbf{RQ3 (Mechanism Breakdown):} How do LLMs process workflow-level demonstrations? Do they rely on semantic memorization of operator names, or do they truly learn the topological routing and execution constraints?
    \item \textbf{RQ4 (Cross-Model Generalization):} Are the synthesized syntactic specifications and topological strategies universal across different LLM foundation models?
\end{itemize}

\begin{table*}[hbtp]
\begin{center}
% \scriptsize
\footnotesize
\setlength{\tabcolsep}{3.5pt}

\caption{Test accuracy (\%) on five benchmarks (GPT-4o-mini). \textbf{Bold} = best, \underline{underline} = runner-up. Baselines from \citet{zhang2025multi}. $^\dagger$Avg.\ over all five. $^\star$MultiArith is \textbf{fully unseen}. $^\S$MaAS uses a different paradigm (probabilistic supernet); shown in gray for reference.}
\label{tab:main_results}
\begin{tabular}{l l cccc c c}
\toprule
& \textbf{Method} & \textbf{GSM8K} & \textbf{MATH} & \textbf{HumanEval} & \textbf{MBPP} & \textbf{MultiArith}$^\star$ & \textbf{Avg.}$^\dagger$ \\
\midrule
\multirow{4}{*}{\rotatebox[origin=c]{90}{\scriptsize\textit{Single}}}
& Vanilla & 87.45 & 46.29 & 87.08 & 71.83 & 96.85 & 77.90 \\
& CoT~\citep{wei2022chain} & 87.10 & 46.40 & 88.13 & 71.83 & 96.31 & 77.95 \\
& ComplexCoT~\citep{fu2022complexity} & 86.89 & 46.53 & 87.49 & 72.36 & 96.70 & 77.99 \\
& SC (CoT$\times$5)~\citep{wang2023selfconsistency} & 87.57 & 47.91 & 88.60 & 73.60 & 96.58 & 78.85 \\
\midrule
\multirow{6}{*}{\rotatebox[origin=c]{90}{\scriptsize\textit{Hand-craft}}}
& MultiPersona~\citep{wang2024unleashing} & 87.50 & 45.43 & 88.32 & 73.19 & 97.49 & 78.39 \\
& LLM-Debate~\citep{du2024improving} & 89.47 & 48.54 & 88.68 & 70.29 & 97.33 & 78.86 \\
& LLM-Blender~\citep{jiang2023llm} & 88.35 & 46.92 & 88.80 & 77.05 & 97.29 & 79.68 \\
& DyLAN~\citep{liu2023dynamic} & 89.98 & 48.63 & 90.42 & 77.30 & 97.12 & 80.69 \\
& AgentVerse~\citep{chen2023agentverse} & 89.91 & 47.35 & 89.29 & 74.28 & 97.50 & 79.67 \\
& MacNet~\citep{qian2024scaling} & 87.95 & 45.18 & 84.57 & 65.28 & 96.03 & 75.80 \\
\midrule
\multirow{6}{*}{\rotatebox[origin=c]{90}{\scriptsize\textit{Automated}}}
& AutoAgents~\citep{chen2023autoagents} & 87.69 & 45.32 & 87.64 & 71.95 & 96.42 & 77.80 \\
& GPTSwarm~\citep{zhuge2024gptswarm} & 89.14 & 47.88 & 89.32 & 77.43 & 96.79 & 80.11 \\
& ADAS~\citep{hu2024automated} & 86.12 & 43.18 & 84.19 & 68.13 & 96.02 & 75.53 \\
& AgentSquare~\citep{shang2024agentsquare} & 87.62 & 48.51 & 89.08 & 78.46 & 97.77 & 80.29 \\

& AFlow~\citep{zhang2025aflow} & 91.16 & 51.28 & 90.93 & 81.67 & 96.22 & 82.25 \\
\rowcolor{gray!15}
& \textcolor{gray}
{MaAS$^\S$~\citep{zhang2025multi}} & \textcolor{gray}{\underline{92.30}} & \textcolor{gray}{\underline{51.82}} & \textcolor{gray}{\underline{92.85}} & \textcolor{gray}{\underline{82.17}} & \textcolor{gray}{\textbf{98.80}} & \textcolor{gray}{83.59} \\
\midrule

& \method{} (ours) & \textbf{94.31} & \textbf{55.14} & \textbf{95.42} & \textbf{83.28} & \underline{98.54} & \textbf{85.34} \\
\bottomrule
\end{tabular}
\end{center}
\vspace{-10pt}
\end{table*}

\paragraph{Datasets \& Leave-One-Out Protocol.} We evaluate \method{} across four in-distribution (ID) and five strictly out-of-distribution (OOD) benchmarks (Table~\ref{tab:dataset_stats} in Appendix~\ref{app:dataset_stats_overall}). To rigorously prevent data contamination and isolate the effect of structural transfer, we employ a strict leave-one-out (LOO) protocol. When synthesizing a solver for a target ID task (e.g., GSM8K), its specific training data, error logs, and previously optimized workflows are entirely masked from the meta-prompt. OOD datasets (e.g., MultiArith, SVAMP, BigCodeBench) remain completely unseen during the trajectory distillation phase, serving as a pure test of structural generalization.

\paragraph{Baselines.} We compare \method{} against: (1) \textbf{Single-agent paradigms} (e.g., CoT~\citep{wei2022chain}, SC~\citep{wang2023selfconsistency}); (2) \textbf{Hand-crafted multi-agent workflows} (e.g., DyLAN~\citep{liu2023dynamic}, AgentVerse~\citep{chen2023agentverse}); and (3) \textbf{Automated test-time search} (e.g., AFlow~\citep{zhang2025aflow}, MaAS~\citep{zhang2025multi}). AFlow serves as our primary baseline, representing the state-of-the-art in iterative and search-based workflow optimization.

\paragraph{Implementation Details.} For our primary evaluations, we use \texttt{gpt-4o-mini} as both the meta-generator $f_\phi$ and the underlying execution engine. To assess cross-model generality, we additionally evaluate \method{} using Grok, Qwen, and Gemma backbones. The trajectory distillation (Phase 1) is performed once entirely offline. Online synthesis (Phase 2) requires exactly one generation pass (temperature $T=0.0$).

\vspace{-3pt}
\subsection{Main Results: Amortized Synthesis Outperforms Iterative Search (RQ1)}
\vspace{-3pt}
\label{subsec:main_results}

Table~\ref{tab:main_results} presents a counter-intuitive result: a single generation pass, with no task-specific search, outperforms the best iteratively optimized workflow. Under our strict leave-one-out protocol, \method{} never observes any workflow code, search trajectory, or error log from the target task; it receives only a brief task description. Yet \method{} consistently outperforms or matches all baselines, including AFlow (the strongest search-based method), across all five benchmarks, at near-zero optimization cost (under \$0.01 per task). This Pareto dominance on both cost and accuracy challenges the assumption that workflow design requires search.

% \textbf{Avoiding the overfitting of test-time search.}
% The gains are largest on benchmarks where search is most prone to overfitting. On GSM8K and HumanEval, \method{} substantially outperforms AFlow (Table~\ref{tab:main_results}) by instantiating robust cognitive architectures such as multi-path sampling with verification for math and test-driven retry for code, transferred from other tasks (Appendix~\ref{app:workflow_examples}), rather than graphs incrementally fine-tuned to each benchmark's validation split.
\textbf{Avoiding the overfitting of test-time search.} The gains are largest on benchmarks where iterative search is inherently prone to overfitting due to small evaluation set sizes (e.g., HumanEval) or highly homogeneous problem structures (e.g., GSM8K). On these datasets, \method{} substantially outperforms AFlow (Table~\ref{tab:main_results}) by instantiating robust, cross-task cognitive architectures such as multi-path sampling with verification for math and test-driven retry for code (Appendix~\ref{app:workflow_examples}), rather than producing idiosyncratic graphs incrementally fine-tuned to a narrow validation split.

\textbf{MultiArith: the purest test of structural transfer.}
MultiArith is entirely absent from the demonstration pool. Yet \method{} achieves $98.54\%$, matching MaAS ($98.80\%$). This provides the strongest evidence that the LLM transfers abstract structural priors, a ``solve then extract'' topology from GSM8K/MATH demos, rather than memorizing dataset-specific heuristics.

\textbf{Differentiated OOD Transfer Across Domains.}
Table~\ref{tab:ood_results} reveals that transfer efficacy varies systematically by domain. Within mathematics, gains are substantial: SVAMP and AQuA show the largest improvements, confirming that the multi-path ensemble topology is domain-invariant, with AQuA showing that \method{} can synthesize multiple-choice extraction steps despite having no multiple-choice workflows in its demonstration pool. AIME gains come primarily from code-assisted verification, but the vast majority of failures reflect the base model's reasoning ceiling rather than workflow limitations (Appendix~\ref{app:ood_failure}). BigCodeBench shows the smallest gain and failure analysis attributes most errors to missing Python dependencies, not workflow topology (Appendix~\ref{app:ood_bcb}).

\begin{wraptable}{r}{0.45\textwidth}
\centering
\vspace{-12pt}
\caption{Out-of-distribution (unseen datasets) generalization results.}
\label{tab:ood_results}
\small
\begin{tabular}{lccc}
\toprule
\textbf{Dataset} & \textbf{Vanilla} & \textbf{\method{}} & \textbf{$\Delta$} \\
\midrule
SVAMP  & 81.0 & \textbf{95.8} & +14.8 \\
AIME 24--25 & 8.3 & \textbf{14.6} & +6.3 \\
AQuA & 52.9 & \textbf{79.9} & +27.0 \\
BigCodeBench & 31.6 & \textbf{34.3} & +2.7 \\
\bottomrule
\end{tabular}
\vspace{-10pt}
\end{wraptable}

\vspace{-3pt}
\subsection{Efficiency: The Structural Smoothness of Workflow Space (RQ2)}
\vspace{-3pt}
\label{subsec:efficiency}
We use MATH as a representative case study, as it is one of the most computationally demanding benchmarks in this paper. \method{} reduces optimization cost by over three orders of magnitude compared to search-based methods (Table~\ref{tab:efficiency_compact}), completing workflow synthesis in under 10 seconds versus $184$ minutes for AFlow. When accounting for the full pipeline (optimization \emph{plus} execution on the test set), \method{} remains $14\times$ cheaper and $29\times$ faster than AFlow, and $2\times$ cheaper and $10\times$ faster than MaAS, while achieving the highest accuracy. The residual cost is almost entirely execution: \method{}'s optimization accounts for less than $0.3\%$ of its total expenditure. In other words, amortized synthesis makes the design of a new workflow essentially free; the only meaningful cost is running it (see Appendix~\ref{app:cost} for a full cost breakdown across all benchmarks). More important than cost reduction is that \method{} Pareto-dominates all search-based methods on \textit{both} cost and accuracy. The fact that eliminating search \textit{improves} accuracy suggests that optimal topologies cluster tightly within domains, and exhaustive graph exploration is not just expensive but actively counterproductive due to validation-set overfitting.

\begin{table}[h!]
\centering
\small
\caption{Efficiency on MATH (GPT-4o-mini). \textbf{Bold} = best, \colorbox{gray!12}{shaded} = runner-up. Multipliers show optimization cost relative to \method{}.}
\label{tab:efficiency_compact}
\renewcommand{\arraystretch}{1.15}
\begin{tabular}{@{}l cc cc cc c@{}}
\toprule
& \multicolumn{2}{c}{\textbf{Optimization}} & \multicolumn{2}{c}{\textbf{Execution}} & \multicolumn{2}{c}{\textbf{Total}} & \\
\cmidrule(lr){2-3} \cmidrule(lr){4-5} \cmidrule(lr){6-7}
\textbf{Method} & Cost (\$) & Time & Cost (\$) & Time & Cost (\$) & Time & Acc.\ (\%) \\
\midrule
LLM-Debate   & ---   & ---   & 6.76 & 92\,min  & 6.76  & 92\,min  & 48.54 \\
DyLAN        & 13.01 & 508\,min & 2.89 & 39\,min  & 15.90 & 547\,min & 48.63 \\
MacNet       & ---   & ---   & 2.35 & 47\,min  & 2.35  & 47\,min  & 45.18 \\
\midrule
GPTSwarm     & 7.02~~{\scriptsize\color{gray}{1755$\times$}}  & 129\,min  & \cellcolor{gray!12}0.93 & 30\,min & 7.95  & 159\,min & 47.88 \\
AFlow        & 22.50~~{\scriptsize\color{red!60!black}{5625$\times$}} & 184\,min  & 1.66 & 23\,min & 24.16~~{\scriptsize\color{red!60!black}{14$\times$}} & 207\,min~~{\scriptsize\color{red!60!black}{30$\times$}} & 51.28 \\
MaAS         & \cellcolor{gray!12}3.38~~{\scriptsize\color{gray}{845$\times$}}   & \cellcolor{gray!12}53\,min   & \textbf{0.42} & \cellcolor{gray!12}19\,min & \cellcolor{gray!12}3.80~~{\scriptsize\color{gray}{2$\times$}}  & \cellcolor{gray!12}72\,min~~{\scriptsize\color{gray}{10$\times$}}  & \cellcolor{gray!12}51.82 \\
\midrule
\rowcolor{green!5}
\rowcolor{green!5}
\textbf{\method{} (Ours)} & \textbf{0.004}\textsuperscript{\dag} & \textbf{0.13\,min}\textsuperscript{\dag} & 1.73 & \textbf{7\,min} & \textbf{1.73} & \textbf{7\,min} & \textbf{55.14} \\
\bottomrule
\multicolumn{8}{@{}l}{\textsuperscript{\dag}\scriptsize Includes one-time contrastive distillation and single-pass optimization.}
% \bottomrule
\end{tabular}
\vspace{-15pt}
\end{table}

\paragraph{The scaling structure of amortized synthesis.}
The cost structures of per-task search and amortized synthesis differ not in degree but in \emph{kind}.
Search-based methods pay the full optimization cost independently for every new task: $n$ tasks cost $n \times C_{\text{search}}$, and no reusable asset is produced.
SWIFT instead converts prior optimization traces into persistent structural priors, reducing each subsequent task to a single generation pass at cost $C_{\text{synth}} \approx \$0.004$.
 
A natural question is whether this comparison should also account for the upstream search trajectories that \method{}'s distillation phase consumes.
Under our leave-one-out protocol, these trajectories are produced exclusively from the optimization of \emph{other} source tasks and are already justified by those tasks' own utility; they would exist regardless of \method{}.
\method{} does not commission additional search for any held-out target; it repurposes existing optimization byproducts at zero marginal search cost.
Even under a deliberately conservative accounting that charges the full source-task search budget ($C_{\text{source}} = \sum_{T \in \mathcal{T}_{\text{src}}} C_{\text{search}}(T) \approx \$112.50$) to SWIFT, the break-even point is approximately five downstream tasks (Appendix~\ref{app:cost_accounting}), after which amortized synthesis is strictly cheaper.
In our evaluation, SWIFT is already applied to nine benchmarks, well beyond this threshold.
 
This analysis highlights a broader point: the leave-one-out protocol is not merely an anti-leakage safeguard.
It is a direct test of whether structural knowledge discovered on one set of tasks can reduce the cost of designing solvers for another.
% The fact that it does, yielding higher accuracy at negligible marginal cost, confirms that per-task search systematically wastes structural knowledge by discarding cross-task regularities that amortized synthesis makes persistent and transferable.
By achieving higher accuracy at near-zero marginal cost, \method{} demonstrates that per-task search is fundamentally wasteful. Instead of discarding shared structural patterns after each task, amortized synthesis turns them into persistent priors that transfer seamlessly to new problems.

\vspace{-3pt}
\subsection{Ablating Meta-Reasoning: Decoupling Syntax and Strategy (RQ3)}
\vspace{-3pt}
\label{subsec:ablations}

\begin{table}[t]
\centering
\caption{Component ablation. $\mathcal{H}$ = compositional heuristics, $\mathcal{C}$ = output contracts. Zero-shot = no demos, Shuffled = randomized code lines, Cross-domain = mismatched-category demos, Random-ops = random operator names. \colorbox{blue!8}{Shaded column}: replacing all operator names with random strings retains $>$93\% average performance, indicating topology, not semantics, drives transfer.}
\label{tab:component_ablation}
\resizebox{\textwidth}{!}{%
\begin{tabular}{l cc cc ccc >{\columncolor{blue!8}}c}
\toprule
& & \multicolumn{2}{c}{\textbf{Component}} & & \multicolumn{4}{c}{\textbf{Demonstration}} \\
\cmidrule(lr){3-4} \cmidrule(lr){6-9}
\textbf{Dataset} & \textbf{Full \method{}} & \textbf{w/o $\mathcal{H}$} & \textbf{w/o $\mathcal{C}$} & & \textbf{Zero-shot} & \textbf{Shuffled} & \textbf{Cross-dom.} & \textbf{Rand-ops} \\
\midrule
GSM8K     & \textbf{94.31} & 88.82\,{\scriptsize\color{red!70!black}$-$5.5} & 89.86\,{\scriptsize\color{red!70!black}$-$4.5} & & 91.19\,{\scriptsize\color{red!70!black}$-$3.1} & 89.38\,{\scriptsize\color{red!70!black}$-$4.9} & 89.29\,{\scriptsize\color{red!70!black}$-$5.0} & 89.67\,{\scriptsize\color{red!70!black}$-$4.6} \\
MATH      & \textbf{55.14} & 49.59\,{\scriptsize\color{red!70!black}$-$5.6} & 50.41\,{\scriptsize\color{red!70!black}$-$4.7} & & \phantom{0}2.47\,{\scriptsize\color{red!70!black}$-$52.7} & \phantom{0}0.00\,{\scriptsize\color{red!70!black}$-$55.1} & 47.74\,{\scriptsize\color{red!70!black}$-$7.4} & 49.59\,{\scriptsize\color{red!70!black}$-$5.6} \\
HumanEval & \textbf{95.42} & 93.13\,{\scriptsize\color{red!70!black}$-$2.3} & 90.84\,{\scriptsize\color{red!70!black}$-$4.6} & & 84.73\,{\scriptsize\color{red!70!black}$-$10.7} & 89.31\,{\scriptsize\color{red!70!black}$-$6.1} & 87.02\,{\scriptsize\color{red!70!black}$-$8.4} & 91.60\,{\scriptsize\color{red!70!black}$-$3.8} \\
MBPP      & \textbf{83.28} & 81.53\,{\scriptsize\color{red!70!black}$-$1.8} & 80.65\,{\scriptsize\color{red!70!black}$-$2.6} & & 72.43\,{\scriptsize\color{red!70!black}$-$10.9} & 69.50\,{\scriptsize\color{red!70!black}$-$13.8} & 70.38\,{\scriptsize\color{red!70!black}$-$12.9} & 77.51\,{\scriptsize\color{red!70!black}$-$5.8} \\
\midrule
Avg.      & \textbf{82.04} & 78.27\,{\scriptsize\color{red!70!black}$-$3.8} & 77.94\,{\scriptsize\color{red!70!black}$-$4.1} & & 62.71\,{\scriptsize\color{red!70!black}$-$19.3} & 62.05\,{\scriptsize\color{red!70!black}$-$20.0} & 73.61\,{\scriptsize\color{red!70!black}$-$8.4} & 77.09\,{\scriptsize\color{red!70!black}$-$4.9} \\
\bottomrule
\end{tabular}%
}
\vspace{-15pt}
\end{table}

To understand \textit{what} the LLM extracts from workflow demonstrations, we perform causal interventions on the meta-prompt (Table~\ref{tab:component_ablation}), revealing a three-level hierarchy.

\begin{wrapfigure}{r}{0.45\textwidth}
\includegraphics[width=1\linewidth]{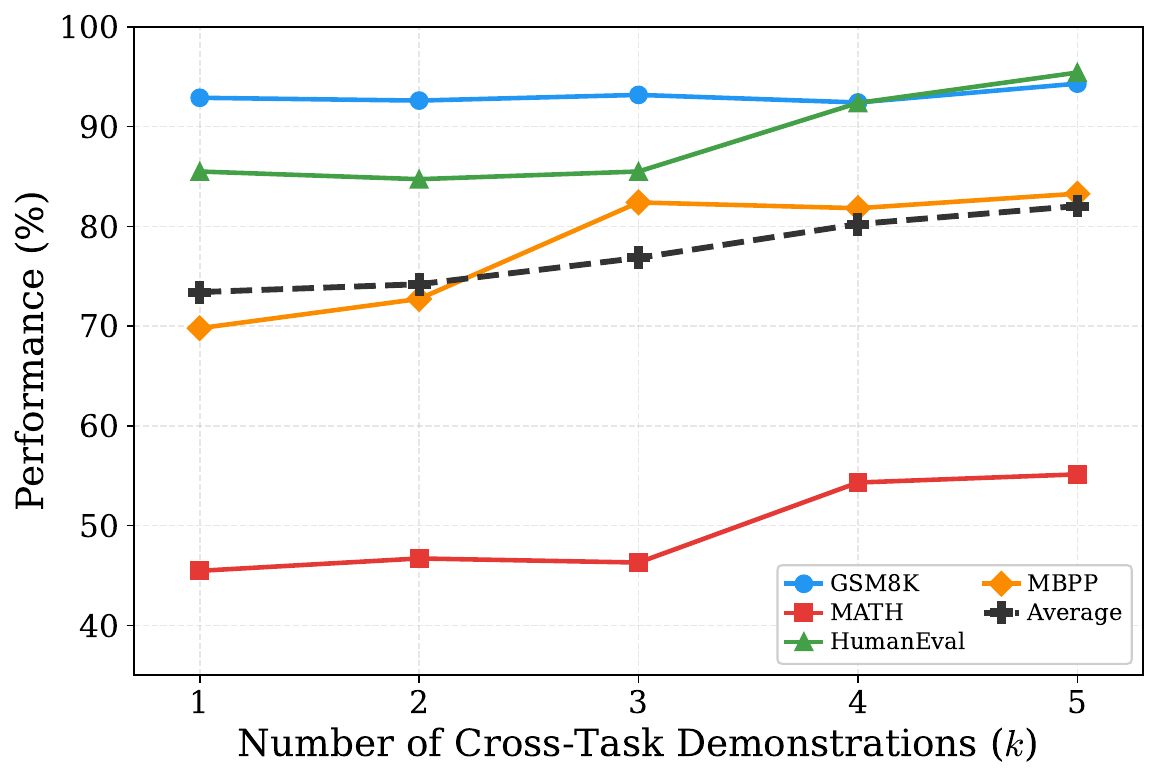}
\vspace{-5pt}
\caption{Effect of demo count ($k$) on performance. GSM8K saturates at $k{=}1$; MATH jumps at $k{=}4$, suggesting complex tasks need more demos to disambiguate workflow topology.}
\label{fig:demo-scaling}
\vspace{-15pt}
\end{wrapfigure}

\textbf{Are demonstrations necessary?}
Removing all demonstrations (\textit{Zero-shot}) collapses MATH to $2.47\%$ yet GSM8K retains $91.19\%$. Simple tasks whose optimal topology is implicit in pretraining need no demos; for complex tasks requiring multi-step workflows with specialized formatting, demos are indispensable. Figure~\ref{fig:demo-scaling} sharpens this finding: GSM8K saturates at $k{=}1$ demonstration while MATH requires $k{=}4$, indicating that complex tasks need multiple demos to disambiguate topology rather than to supply additional semantic content.

\textbf{Topology, not semantics.}
The \textit{Random-ops} intervention yields the most surprising finding: replacing all operator names with random strings barely degrades performance---less than $5$ points on any benchmark (Table~\ref{tab:component_ablation}). The LLM learns \textit{topological routing}, directed edges and data flow, independent of human-readable labels.

\textbf{Domain boundaries.}
Mismatched-domain demonstrations (\textit{Cross-domain}) cause targeted degradation: MATH and MBPP degrade substantially (Table~\ref{tab:component_ablation}). Code feedback loops (generate $\to$ test $\to$ retry) are topologically incompatible with math verification (multi-path $\to$ ensemble $\to$ extract). Optimal topologies are \textit{domain-specific but task-invariant}: convergent within a family, distinct across families.

\textbf{Bridging the structural-functional gap.}
Removing output contracts ($-$$\mathcal{C}$) substantially degrades accuracy (Table~\ref{tab:component_ablation}). The meta-generator designs logically sound workflows but terminal nodes output conversational text violating exact-match parsers. Contracts ($\mathcal{C}$) act as a type system that ensures workflow synthesis is a compilation task with strict interface guarantees. A related failure mode is that \texttt{ScEnsemble} fundamentally fails for code because functionally equivalent programs are never textually identical (Appendix~\ref{sec:negative_results}).

\vspace{-3pt}
\subsection{Generality Across Model Families (RQ4)}
\vspace{-3pt}
\label{subsec:generality}

To ensure the structural priors synthesized by \method{} are not overfitted artifacts of OpenAI's alignment tax, we apply \method{}-generated workflows using open-weight and alternative proprietary backbones as the execution engine (Table~\ref{tab:generality}). Across Grok-4.1-fast, Qwen3-8B, and Gemma-3-12B-IT, \method{} consistently improves accuracy on 16 of 18 benchmark-model pairs, with particularly large gains on MATH and MBPP (Table~\ref{tab:generality}). This suggests that the topological strategies produced via \method{} act as universal cognitive scaffolding that unlocks latent reasoning capabilities regardless of the underlying foundation model.

One notable exception reveals an important boundary condition: on Gemma-3-12B-IT, MATH accuracy \textit{decreases} from $58.23\%$ to $48.35\%$. The MATH workflow chains multi-path reasoning, ensemble voting, and final-answer parsing across multiple operators, placing high demands on instruction-following fidelity. We hypothesize that Gemma's instruction-following capability is insufficient for such complex workflows, causing error accumulation across nodes. This suggests structural transfer requires a minimum instruction-following ability of the base model.

\begin{table}[h!]
\centering
\caption{Generality across model families. Bold = best per backbone-benchmark pair.}
\label{tab:generality}
\resizebox{\textwidth}{!}{%
\begin{tabular}{llcccccc|c}
\toprule
\textbf{Backbone} & \textbf{Method} & \textbf{GSM8K} & \textbf{MATH} & \textbf{HumanEval} & \textbf{MBPP} & \textbf{DROP} & \textbf{HotpotQA} & \textbf{Avg.} \\
\midrule
\multirow{3}{*}{Grok-4.1-fast}
 & Vanilla            & 92.23 & 44.03 & 87.02 & 64.29 & 68.91 & 55.15 & 68.61 \\
 & \method{} (Ours)   & \textbf{93.93} & \textbf{59.05} & \textbf{91.60} & \textbf{85.30} & \textbf{71.17} & \textbf{78.00} & \textbf{79.84} \\
 & $\Delta$           & \cellcolor{green!10}{+1.70} & \cellcolor{green!10}{+15.02} & \cellcolor{green!10}{+4.58} & \cellcolor{green!10}{+21.01} & \cellcolor{green!10}{+2.26} & \cellcolor{green!10}{+22.85} & \cellcolor{green!10}{+11.24} \\
\midrule
\multirow{3}{*}{Qwen3-8B}
 & Vanilla            & 88.44 & 61.11 & 85.50 & 66.28 & 49.69 & 65.49 & 69.42 \\
 & \method{} (Ours)   & \textbf{92.51} & \textbf{66.46} & \textbf{88.55} & \textbf{72.73} & \textbf{83.21} & \textbf{72.68} & \textbf{79.36} \\
 & $\Delta$           & \cellcolor{green!10}{+4.07} & \cellcolor{green!10}{+5.35} & \cellcolor{green!10}{+3.05} & \cellcolor{green!10}{+6.45} & \cellcolor{green!10}{+33.52} & \cellcolor{green!10}{+7.19} & \cellcolor{green!10}{+9.94} \\
\midrule
\multirow{3}{*}{Gemma-3-12B-IT}
 & Vanilla            & 91.85 & \textbf{58.23} & 84.73 & 71.55 & \textbf{63.13} & 71.19 & 73.45 \\
 & \method{} (Ours)   & \textbf{93.46} & 48.35 & \textbf{87.02} & \textbf{75.95} & 63.00 & \textbf{73.36} & \textbf{73.52} \\
 & $\Delta$           & \cellcolor{green!10}{+1.61} & \cellcolor{red!10}{$-$9.88} & \cellcolor{green!10}{+2.29} & \cellcolor{green!10}{+4.40} & \cellcolor{red!10}{$-$0.13} & \cellcolor{green!10}{+2.17} & \cellcolor{green!10}{+0.08} \\
\bottomrule
\end{tabular}%
}
% \vspace{-15pt}
\end{table}

% Failure Mode Analysis is covered in Appendices B and C
\section{Discussion}
% \vspace{-8pt}
The empirical success of \method{} invites a reevaluation of how the field currently approaches autonomous agent design. By transitioning from test-time combinatorial search to amortized synthesis guided by demonstrations, our findings highlight several broader implications for LLM workflows.
% \vspace{-3pt}

\textbf{The Amortization of Search Costs.}
While acquiring high-quality demonstrations inherently requires an initial upstream search budget, our results suggest that MCTS-based workflow optimization often searches a functionally redundant space. Table~\ref{tab:efficiency_compact} quantifies the waste: AFlow invests over $5{,}000\times$ more optimization compute yet achieves lower accuracy than \method{}'s single-pass synthesis. Optimal topologies are highly convergent within domains (Appendix~\ref{app:workflow_examples}).
\vspace{-3pt}

\textbf{The Necessity of Contract-Guided Generation.}
As LLMs grow more capable, the bottleneck shifts from reasoning to interfacing. Our contract ablation (Table~\ref{tab:component_ablation}) shows that agents can possess correct cognitive topologies yet fail due to output formatting violations. Workflow synthesis must be treated as a compilation task with strict type-safety guarantees.
% \vspace{-3pt}

\paragraph{Limitations and Future Work.}
While SWIFT effectively solves static reasoning and generation benchmarks, real-world agentic deployment is rarely static. Future work can extend amortized synthesis to \emph{stochastic and irreversible execution environments}, where actions may alter environment state permanently. Amortized synthesis is also not a universal free lunch, and our OOD analysis (Appendix~\ref{app:ood_failure}) surfaces four regimes where it breaks down.

First, when failures are \emph{capability-bounded}: on AIME 24--25, SWIFT plateaus at $14.6\%$ ($+6.3$ over vanilla, Table~\ref{tab:ood_results}), indicating a reasoning ceiling that little topological scaffolding can lift. Second, when failures are \emph{environment-bounded}: $71\%$ of BigCodeBench failures
are missing-module errors (Figure~\ref{fig:bcb_breakdown}), leaving a true logic-error rate comparable to MBPP---the sandbox, not the workflow, is the bottleneck. Third, when the target task's reasoning pattern is \emph{absent from the source pool}: on non-arithmetic AQuA puzzles (e.g., the letter-counting ``mango'' case in Appendix~\ref{app:ood_aqua}), the
math-focused strategy transferred from GSM8K/MATH actively misleads the synthesized workflow toward numerical computation, suggesting transfer can degrade rather than merely under-deliver when source and target strategies mismatch. Fourth, when the execution model's instruction-following is insufficient for the synthesized topology: Gemma-3-12B-IT's MATH accuracy drops from $58.23\% \to 48.35\%$ under SWIFT (Table~\ref{tab:generality}),
as chained multi-path reasoning, ensemble voting, and format extraction accumulate instruction-following errors across nodes. Understanding and quantifying this minimum capability threshold remains an open question, and more broadly, predicting transfer success from task--pool similarity before synthesis would turn these current regimes into an actionable criterion.
Furthermore, replacing our heuristically distilled contracts with rigorous formal methods could guarantee that a synthesized workflow adheres to safety and functional constraints prior to execution.

% \enlargethispage{6\baselineskip}
% \vspace{-3pt}
\section{Conclusion}
% \vspace{-8pt}
\label{sec:conclusion}
In this work, we introduced \method{}, a framework that reframes LLM workflow optimization from costly, iterative test-time search to amortized synthesis guided by demonstrations. By conditioning a meta-generator on cross-task programmatic demonstrations, compositional heuristics, and output contracts, \method{} deduces optimal solver architectures in a single generation pass. Empirically, \method{} not only matches or exceeds state-of-the-art search baselines like AFlow across 5 benchmarks, but does so while reducing optimization costs by three orders of magnitude. Crucially, our mechanism analyses reveal that LLMs possess the ability to transfer abstract, topological routing strategies across task boundaries, independent of semantic labels. By explicitly bridging the structural-functional gap through contract-guided generation, \method{} provides a highly efficient, scalable way for the automated design of autonomous AI agents.

\section*{Acknowledgment}
This work was supported in part by the US National Science Foundation [III-2232121] and the US National Institutes of Health [R01HG012470]. Shiyi Du was partially supported by the SoftBank Group / Arm Ph.D. Fellowship. Jiayuan Liu and Vincent Conitzer thank the Cooperative AI Foundation,  Macroscopic Ventures (formerly Polaris Ventures / the Center for Emerging Risk Research) and Jaan Tallinn’s donor-advised fund at Founders Pledge for financial support. Weihua Du was supported in part by SoftBank Group Corp. and Arm, Inc. Yue Huang was supported by a JetStream2 Fellowship. Jiayi Li was partially supported by a gift from Arm, Inc. Conflict of Interest: Carl Kingsford is a co-founder of Ellumigen, Inc.

% \newpage
\bibliography{colm2026_conference}
\bibliographystyle{colm2026_conference}

\newpage
\appendix

\section{Dataset Statistics}
\label{app:dataset_stats_overall}
\begin{table}[H]
\centering
\caption{Dataset statistics. \textit{Unseen} denotes datasets not included in the demonstration pool during workflow generation (i.e., purely out-of-distribution). \textsuperscript{\dag}AIME combines AIME 2024 and AIME 2025.}
\label{tab:dataset_stats}
\begin{tabular}{llcrrc}
\toprule
\textbf{Dataset} & \textbf{Task Type} & \textbf{Test} & \textbf{Val} & \textbf{Unseen} \\
\midrule
\multicolumn{5}{l}{\textit{In-Distribution (used as leave-one-out demonstrations)}} \\
\midrule
GSM8K      & Math (Elementary)       & 1{,}055 & 264 &  \\
MATH       & Math (Competition)      & 486     & 119 &  \\
HumanEval  & Code Generation         & 131     & 33  &  \\
MBPP       & Code Generation         & 341     & 86  &  \\
\midrule
\multicolumn{5}{l}{\textit{Out-of-Distribution (entirely absent from demonstration pool)}} \\
\midrule
MultiArith    & Math (Elementary)       & 480  & 120 & \checkmark \\
SVAMP         & Math (Elementary)       & 800  & 200 & \checkmark \\
AIME\textsuperscript{\dag}          & Math (Competition)      & 48   & 12  & \checkmark \\
AQUA          & Math (Multiple Choice)  & 204  & 50  & \checkmark \\
BigCodeBench  & Code Generation         & 912  & 228 & \checkmark \\
\midrule
\textbf{Total} &                        & \textbf{4{,}457} & \textbf{1{,}112} & \\
\bottomrule
\end{tabular}
\end{table}

\section{Algorithm}
\label{app:algorithm}

\begin{algorithm}[htbp]
\caption{\methodfull{} (\method{})}
\label{alg:amortized_synthesis}
\begin{algorithmic}[1]
\REQUIRE All available tasks $\mathcal{T}_{all}$, Target task $\mathcal{T}_{target}$, Operator library $\mathcal{O}$.
\REQUIRE Historical trajectories $\{ \tau_{\mathcal{T}} \}_{\mathcal{T} \in \mathcal{T}_{all}}$, Threshold $\gamma$.
\REQUIRE Meta-generator LLM $f_\phi$, Contrastive reflection LLM $f_\psi$.
\vspace{1mm}
\STATE \textbf{\% Definition:} Source tasks $\mathcal{T}_{src} \leftarrow \mathcal{T}_{all} \setminus \{\mathcal{T}_{target}\}$ \hfill \COMMENT{Strict Leave-One-Out constraint}
\vspace{1mm}
\STATE \textbf{\% Phase 1: Contrastive Trajectory Distillation (Offline)}
\STATE Initialize global heuristics $\mathcal{H} \leftarrow \emptyset$, output contracts $\mathcal{C} \leftarrow \emptyset$
\FOR{each task $\mathcal{T} \in \mathcal{T}_{src}$}
    \STATE $\tau \leftarrow \tau_{\mathcal{T}}$
    \STATE $w_{best} \leftarrow \arg\max_{w \in \tau} \text{Acc}(w)$ \hfill \COMMENT{Isolate the optimal structural topology}
    \STATE $\tau_{low} \leftarrow \{ w \in \tau \mid 0 < \text{Acc}(w) < \gamma \cdot \text{Acc}(w_{best}) \}$ \hfill \COMMENT{Low-scoring workflows ($\gamma{=}0.6$)}
    \STATE $\tau_{zero} \leftarrow \{ w \in \tau \mid \text{Acc}(w) = 0 \}$ \hfill \COMMENT{Collapsed workflows}
    \STATE $w_{low} \leftarrow \arg\min_{w \in \tau_{low}} \text{Acc}(w)$; \quad $w_{zero} \leftarrow \text{first}(\tau_{zero})$ \hfill \COMMENT{Hard negatives}
    \STATE Extract execution traces and parsing error logs $E_{low}, E_{zero}$ from $w_{low}, w_{zero}$
    \STATE \textbf{\% Distill topological priors and functional contracts via contrastive reflection}
    \STATE $\mathcal{H}_{\mathcal{T}}, \mathcal{C}_{\mathcal{T}} \leftarrow f_\psi(w_{best}, w_{low}, E_{low}, w_{zero}, E_{zero})$
    \STATE $\mathcal{H} \leftarrow \mathcal{H} \cup \mathcal{H}_{\mathcal{T}}$, \quad $\mathcal{C} \leftarrow \mathcal{C} \cup \mathcal{C}_{\mathcal{T}}$
\ENDFOR
\vspace{1mm}
\STATE \textbf{\% Phase 2: Few-Shot Amortized Synthesis (Online)}
\STATE \textbf{\% Retrieve programmatic demonstrations exclusively from source tasks (OOD)}
\STATE $\mathcal{D}_{demo} \subset \{ w_{best}^{(\mathcal{T})} \mid \mathcal{T} \in \mathcal{T}_{src} \}$
\STATE Construct meta-prompt $P = (\mathcal{O}, \mathcal{H}, \mathcal{C}, \mathcal{D}_{demo}, \mathcal{T}_{target})$
\vspace{1mm}
\STATE \textbf{\% End-to-end Contract-Guided Generation (Single generation pass)}
\STATE $w_{target} \leftarrow f_\phi(P)$ \hfill \COMMENT{Workflow natively adheres to $\mathcal{H}$ and $\mathcal{C}$}
\RETURN Executable workflow $w_{target}$
\end{algorithmic}
\end{algorithm}

\section{Generated Workflow Examples}
\label{app:workflow_examples}

% This section presents the complete workflow code synthesized
Below we show the codes of the workflows synthesized by \method{}.
All workflows are generated in a single generation pass from leave-one-out demonstrations. Boilerplate (imports, \texttt{\_\_init\_\_}) is omitted for clarity.

\paragraph{GSM8K.}
Three independent reasoning paths are generated and aggregated via \textsc{ScEnsemble} (self-consistency voting), followed by a dedicated extraction step.

\begin{lstlisting}[style=workflow, caption={GSM8K workflow.}]
class Workflow:
    def __init__(self, name, llm_config, dataset):
        self.name = name
        self.dataset = dataset
        self.llm = create_llm_instance(llm_config)

        # Operators
        self.custom = Custom(self.llm)
        self.sc_ensemble = ScEnsemble(self.llm)

    async def __call__(self, problem: str):
        # Step 1: Generate 3 independent reasoning paths
        solutions = []
        for _ in range(3):
            reasoning = await self.custom(
                input=problem,
                instruction=(
                    "Break down the problem step by step, "
                    "extract the necessary numbers, perform "
                    "the calculations, and provide only the "
                    "final answer as a single number."
                )
            )
            solutions.append(reasoning['response'])

        # Step 2: Self-consistency ensemble voting
        ensemble_result = await self.sc_ensemble(
            solutions=solutions,
            problem=problem
        )

        # Step 3: Extract clean numerical answer
        final = await self.custom(
            input=(
                f"Final answer based on reasoning: "
                f"{ensemble_result['response']}"
            ),
            instruction=(
                "Extract ONLY the final numerical answer. "
                "No units, no explanation."
            )
        )

        return final['response'], self.llm.cost_manager.total_cost
\end{lstlisting}

\paragraph{MATH.}
Same multi-path + ensemble structure as GSM8K, with a \LaTeX{} \verb|\boxed{}| extraction step tailored to the MATH answer format

\begin{lstlisting}[style=workflow, caption={MATH workflow.}]
class Workflow:
    def __init__(self, name, llm_config, dataset):
        self.name = name
        self.dataset = dataset
        self.llm = create_llm_instance(llm_config)

        # Operators
        self.custom = Custom(self.llm)
        self.sc_ensemble = ScEnsemble(self.llm)

    async def __call__(self, problem: str):
        # Step 1: Generate 3 independent solutions
        solutions = []
        for _ in range(3):
            solution = await self.custom(
                input=problem,
                instruction=(
                    "Solve the problem step by step "
                    "and show all calculations."
                )
            )
            solutions.append(solution['response'])

        # Step 2: Self-consistency ensemble voting
        final_solution = await self.sc_ensemble(
            solutions=solutions,
            problem=problem
        )

        # Step 3: Extract answer in LaTeX \boxed{} format
        formatted_answer = await self.custom(
            input=(
                f"Final answer: "
                f"{final_solution['response']}"
            ),
            instruction=(
                "Extract ONLY the final answer "
                "in \\boxed{} format."
            )
        )

        return (
            formatted_answer['response'],
            self.llm.cost_manager.total_cost
        )
\end{lstlisting}

\paragraph{HumanEval.}
A test-driven retry loop: generate code, run unit tests, and if the tests fail, analyze the error and regenerate, up to two repair attempts.

\begin{lstlisting}[style=workflow, caption={HumanEval workflow.}]
class Workflow:
    def __init__(self, name, llm_config, dataset):
        self.name = name
        self.dataset = dataset
        self.llm = create_llm_instance(llm_config)

        # Operators
        self.custom = Custom(self.llm)
        self.custom_code_generate = CustomCodeGenerate(self.llm)
        self.test = Test(self.llm, dataset=self.dataset)

    async def __call__(self, problem: str, entry_point: str):
        # ---- Attempt 1: initial generation ----
        initial_solution = await self.custom_code_generate(
            problem=problem,
            entry_point=entry_point,
            instruction=(
                "Generate a Python function that meets "
                "the specifications in the docstring."
            )
        )

        test_result = await self.test(
            problem=problem,
            solution=initial_solution['response'],
            entry_point=entry_point
        )

        if test_result['result']:
            return (
                initial_solution['response'],
                self.llm.cost_manager.total_cost
            )

        # ---- Attempt 2: analyze failure and fix ----
        analysis = await self.custom(
            input=(
                f"Problem: {problem}\n"
                f"Failed code: "
                f"{initial_solution['response']}\n"
                f"Error: {test_result['solution']}"
            ),
            instruction=(
                "Analyze why this code failed "
                "and how to fix it"
            )
        )

        improved = await self.custom_code_generate(
            problem=problem,
            entry_point=entry_point,
            instruction=(
                f"Fix the code based on: "
                f"{analysis['response']}"
            )
        )

        test_result2 = await self.test(
            problem=problem,
            solution=improved['response'],
            entry_point=entry_point
        )

        if test_result2['result']:
            return (
                improved['response'],
                self.llm.cost_manager.total_cost
            )

        # ---- Attempt 3: regenerate from scratch ----
        final = await self.custom_code_generate(
            problem=problem,
            entry_point=entry_point,
            instruction=(
                f"Previous attempts failed. "
                f"Error: {test_result2['solution']}. "
                f"Generate a correct solution."
            )
        )

        return (
            final['response'],
            self.llm.cost_manager.total_cost
        )
\end{lstlisting}

\paragraph{MultiArith.}
A simple two-step workflow: detailed step-by-step reasoning followed by numerical extraction.
Despite having no MultiArith demonstrations, \method{} synthesizes a clean ``solve then extract'' pattern transferred from the math-reasoning source demos.

\begin{lstlisting}[style=workflow, caption={MultiArith workflow.}]
class Workflow:
    def __init__(self, name, llm_config, dataset):
        self.name = name
        self.dataset = dataset
        self.llm = create_llm_instance(llm_config)

        # Operators
        self.custom = Custom(self.llm)

    async def __call__(self, problem: str):
        # Step 1: Detailed step-by-step reasoning
        detailed_response = await self.custom(
            input=problem,
            instruction=(
                "You are a skilled mathematician. Please "
                "solve the following arithmetic word "
                "problem step-by-step:\n"
                "1. Read and extract all relevant "
                "numerical information and operations.\n"
                "2. Clearly state the mathematical "
                "operations in a logical sequence.\n"
                "3. Perform the calculations step-by-step, "
                "showing your work clearly.\n"
                "4. Provide a concise final answer as a "
                "single numerical value without any "
                "additional text or units."
            )
        )

        # Step 2: Extract clean numerical answer
        final_answer = await self.custom(
            input=detailed_response['response'],
            instruction=(
                "Given the detailed calculations and the "
                "answer derived from the arithmetic "
                "problem above, please provide ONLY the "
                "numerical answer without any additional "
                "text or units."
            )
        )

        return (
            final_answer['response'],
            self.llm.cost_manager.total_cost
        )
\end{lstlisting}

\paragraph{AIME.}
A hybrid workflow combining \textsc{Programmer} (code-assisted computation) with multi-path reasoning and ensemble.
This is the most complex OOD workflow, reflecting the difficulty of competition-level math.

\begin{lstlisting}[style=workflow, caption={AIME workflow (14.58\%).}]
class Workflow:
    def __init__(self, name, llm_config, dataset):
        self.name = name
        self.dataset = dataset
        self.llm = create_llm_instance(llm_config)

        # Operators
        self.custom = Custom(self.llm)
        self.programmer = Programmer(self.llm)
        self.sc_ensemble = ScEnsemble(self.llm)

    async def __call__(self, problem: str):
        # Step 1: Code-assisted computation
        code_solution = await self.programmer(
            problem=problem
        )

        # Step 2: Integrate code output with reasoning
        formatted_solution = await self.custom(
            input=(
                problem + f"\nCode output: "
                f"{code_solution['output']}"
            ),
            instruction=(
                "Given the problem and code output, "
                "provide a detailed solution with LaTeX. "
                "Show step-by-step calculations. "
                "Present the final answer in "
                "\\boxed{} notation."
            )
        )

        # Step 3: Generate 3 additional reasoning paths
        solutions = [formatted_solution['response']]
        for _ in range(3):
            alternative = await self.custom(
                input=problem,
                instruction=(
                    "Solve step by step with LaTeX. "
                    "Outline the approach and relevant "
                    "formulas. Provide detailed "
                    "calculations. Present the final "
                    "answer in \\boxed{} notation."
                )
            )
            solutions.append(alternative['response'])

        # Step 4: Self-consistency ensemble voting
        final_solution = await self.sc_ensemble(
            solutions=solutions,
            problem=problem
        )

        # Step 5: Extract final answer in \boxed{} format
        refined = await self.custom(
            input=final_solution['response'],
            instruction=(
                "Extract only the final answer. "
                "Respond strictly in the format "
                "\\boxed{answer}, without any "
                "additional text or explanation."
            )
        )

        return (
            refined['response'],
            self.llm.cost_manager.total_cost
        )
\end{lstlisting}

\paragraph{BigCodeBench.}
Three candidate implementations voted via \textsc{ScEnsemble}, validated by unit tests, with a single retry on failure.
The test-driven pattern is transferred from the HumanEval and MBPP source demos.

\begin{lstlisting}[style=workflow, caption={BigCodeBench workflow.}]
class Workflow:
    def __init__(self, name, llm_config, dataset):
        self.name = name
        self.dataset = dataset
        self.llm = create_llm_instance(llm_config)

        # Operators
        self.custom_code_generate = CustomCodeGenerate(
            self.llm
        )
        self.sc_ensemble = ScEnsemble(self.llm)
        self.test = Test(self.llm, dataset=self.dataset)

    async def __call__(self, problem: str, entry_point: str):
        # Step 1: Generate 3 candidate implementations
        solutions = []
        for _ in range(3):
            solution = await self.custom_code_generate(
                problem=problem,
                entry_point=entry_point,
                instruction=""
            )
            solutions.append(solution['response'])

        # Step 2: Ensemble voting to select best candidate
        best_solution = await self.sc_ensemble(
            solutions=solutions,
            problem=problem
        )

        # Step 3: Validate with unit tests
        test_result = await self.test(
            problem=problem,
            solution=best_solution['response'],
            entry_point=entry_point
        )

        if test_result['result']:
            return (
                test_result['solution'],
                self.llm.cost_manager.total_cost
            )

        # Step 4: Single retry with error context
        improved = await self.custom_code_generate(
            problem=problem,
            entry_point=entry_point,
            instruction=(
                "The previous solution did not pass "
                "the test cases. Analyze requirements "
                "carefully and generate a new correct "
                "solution."
            )
        )

        return (
            improved['response'],
            self.llm.cost_manager.total_cost
        )
\end{lstlisting}

\section{ScEnsemble fails on code generation tasks}
\label{sec:negative_results}
\label{sec:neg_scensemble_code}

An important early finding was that \texttt{ScEnsemble} (self-consistency ensemble via text-based majority voting) \textbf{does not work for code generation tasks}. \texttt{ScEnsemble} selects the most frequent answer among multiple sampled solutions by comparing their textual representations. This is effective for math and QA tasks where the final answer is a short, canonical string (e.g., ``42'', ``Paris''), but fundamentally breaks for code because:

\begin{enumerate}[nosep, leftmargin=*]
    \item \textbf{Surface-form diversity.} Functionally equivalent programs can differ vastly in variable names, control flow structure, whitespace, and style. Two correct implementations of the same function will almost never be textually identical.
    \item \textbf{No semantic equivalence check.} Text-based voting treats each character difference as a ``disagreement,'' so even trivially equivalent programs (e.g., \texttt{for i in range(n)} vs.\ \texttt{for i in range(0, n)}) are counted as distinct solutions.
    \item \textbf{Chimera outputs.} When no clear majority exists, the ensemble prompt may produce a ``compromise'' snippet that combines fragments from multiple solutions, resulting in code that does not compile or execute.
\end{enumerate}

\noindent\textbf{Observation.} Across our experiments, we found that synthesized workflows which rely solely on \texttt{ScEnsemble} for code tasks consistently underperform those that incorporate a \textbf{test-driven retry} pattern (generate code $\rightarrow$ execute against test cases $\rightarrow$ if tests fail, regenerate with error feedback). Execution-based validation assesses \emph{functional} correctness rather than surface-form similarity.

This finding also generalizes beyond \method{}: any workflow system that applies self-consistency voting as the sole correctness criterion for code will encounter the same limitation. Execution-based evaluation is necessary for tasks where output correctness is defined by behavior rather than surface form.

\textbf{Clarification.} Some synthesized workflows (e.g., BigCodeBench, Supplementary Section~\ref{app:workflow_examples}) still include \texttt{ScEnsemble} as a pre-filtering step before test execution. This is not contradictory: when ensemble voting serves only as a heuristic candidate selector---with execution-based testing as the final arbiter of correctness---the surface-form equivalence problem is mitigated. The failure mode we identify here is using text-based voting as the \emph{terminal} correctness check with no subsequent execution validation.

\section{Out-of-distribution Failure Analysis}
\label{app:ood_failure}

\begin{figure}[t]
\centering
\begin{tikzpicture}
\definecolor{correct}{HTML}{27AE60}   % green - correct
\definecolor{modelerr}{HTML}{E74C3C}  % red - exec model error
\definecolor{wferr}{HTML}{F39C12}     % orange - workflow design error
\definecolor{enverr}{HTML}{3498DB}    % blue - infrastructure/env error
\definecolor{vanilla}{HTML}{BDC3C7}   % gray - vanilla baseline

\def\bw{0.38}
\def\ml{8.0}
\def\gap{0.18}

% --- SVAMP ---
\node[anchor=east, font=\small\bfseries] at (-0.15, 4.9) {SVAMP};

\fill[correct, rounded corners=1pt] (0, 4.9-0.12) rectangle ({\ml*0.958}, 4.9+0.12);
\fill[modelerr, rounded corners=1pt] ({\ml*0.958}, 4.9-0.12) rectangle ({\ml*1.0}, 4.9+0.12);
\node[anchor=west, font=\scriptsize] at ({\ml+0.08}, 4.9) {\textbf{95.8}};

% --- AQuA ---
\node[anchor=east, font=\small\bfseries] at (-0.15, 3.5) {AQuA};

\fill[correct, rounded corners=1pt] (0, 3.5-0.12) rectangle ({\ml*0.799}, 3.5+0.12);
\fill[modelerr, rounded corners=1pt] ({\ml*0.799}, 3.5-0.12) rectangle ({\ml*0.988}, 3.5+0.12);
\fill[wferr, rounded corners=1pt]    ({\ml*0.988}, 3.5-0.12) rectangle ({\ml*1.0}, 3.5+0.12);
\node[anchor=west, font=\scriptsize] at ({\ml+0.08}, 3.5) {\textbf{79.9}};

% --- AIME ---
\node[anchor=east, font=\small\bfseries] at (-0.15, 2.1) {AIME};

\fill[correct, rounded corners=1pt] (0, 2.1-0.12) rectangle ({\ml*0.146}, 2.1+0.12);
\fill[modelerr, rounded corners=1pt] ({\ml*0.146}, 2.1-0.12) rectangle ({\ml*1.0}, 2.1+0.12);
\node[anchor=west, font=\scriptsize] at ({\ml+0.08}, 2.1) {\textbf{14.6}};

% --- BigCodeBench ---
\node[anchor=east, font=\small\bfseries] at (-0.15, 0.7) {BigCodeBench};
\fill[correct, rounded corners=1pt] (0, 0.7-0.12) rectangle ({\ml*0.343}, 0.7+0.12);
\fill[modelerr, rounded corners=1pt] ({\ml*0.343}, 0.7-0.12) rectangle ({\ml*0.290+\ml*0.343-\ml*0.343+\ml*0.117}, 0.7+0.12);
\fill[correct, rounded corners=1pt]  (0, 0.7-0.12)           rectangle ({\ml*0.343}, 0.7+0.12);
\fill[modelerr, rounded corners=1pt] ({\ml*0.343}, 0.7-0.12) rectangle ({\ml*0.460}, 0.7+0.12);
\fill[wferr, rounded corners=1pt]    ({\ml*0.460}, 0.7-0.12) rectangle ({\ml*0.578}, 0.7+0.12);
\fill[enverr, rounded corners=1pt]   ({\ml*0.578}, 0.7-0.12) rectangle ({\ml*1.0}, 0.7+0.12);
\node[anchor=west, font=\scriptsize] at ({\ml+0.08}, 0.7) {\textbf{34.3}};

% --- Legend ---
\node[anchor=west, font=\scriptsize] at (0, -0.4) {
    \tikz{\fill[correct, rounded corners=1pt] (0,0) rectangle (0.25,0.18);}\ Correct
    \hspace{4pt}
    \tikz{\fill[modelerr, rounded corners=1pt] (0,0) rectangle (0.25,0.18);}\ Model error
    \hspace{4pt}
    \tikz{\fill[wferr, rounded corners=1pt] (0,0) rectangle (0.25,0.18);}\ Workflow issue
    \hspace{4pt}
    \tikz{\fill[enverr, rounded corners=1pt] (0,0) rectangle (0.25,0.18);}\ Env issue
    \hspace{4pt}
    \tikz{\fill[vanilla, rounded corners=1pt] (0,0) rectangle (0.25,0.18);}\ Vanilla baseline
};

\end{tikzpicture}
\caption{OOD performance overview. Each \method{} bar decomposes the full test set into correct (green), execution model errors (red), workflow design issues (orange), and infrastructure errors (blue). AIME is dominated by model errors (competition-level math). BigCodeBench is the only benchmark where infrastructure issues (missing Python modules) constitute the largest failure category.}
\label{fig:ood_perf_overview}
\end{figure}
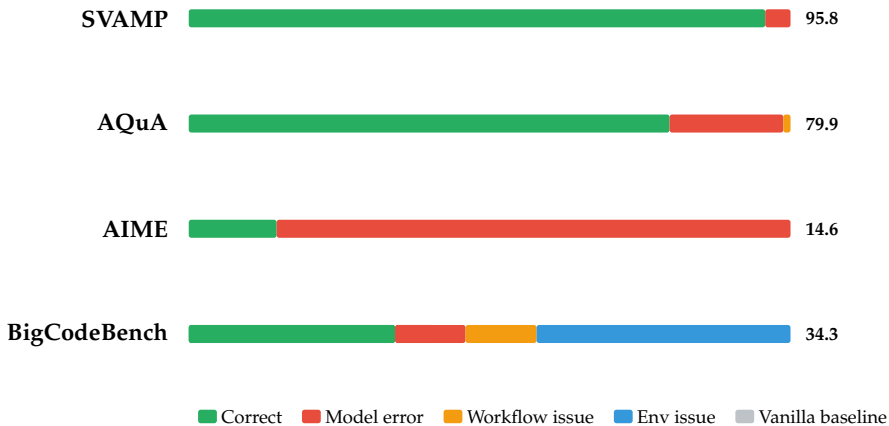

We analyze failure modes on four out-of-distribution (OOD) benchmarks---datasets \emph{entirely unseen} during demonstration construction.
\method{} synthesizes workflows for these datasets using only in-distribution demonstrations, making failures particularly informative: they reveal where cross-task transfer succeeds and where it breaks down. Figure~\ref{fig:ood_perf_overview} summarizes the OOD performance overview.

% ======================================================
% FIGURE 1: Performance overview with error breakdown
% ======================================================

\subsection{SVAMP: adversarial word problems}
\label{app:ood_svamp}

\noindent
\textbf{Score}: 95.8\% (34 failures / 800 problems). \textbf{Gain over vanilla}: +14.8 points. \\
\textbf{Workflow}: 5$\times$Custom (solve) $\to$ ScEnsemble (vote) $\to$ Programmer (code verification).

SVAMP is designed to test robustness against surface-form variation of grade-school math problems.
The gain over vanilla GPT-4o-mini demonstrates strong cross-task transfer from GSM8K/MATH demonstrations: the synthesized ensemble+verification pipeline provides systematic error correction that vanilla prompting lacks.

The 34 remaining failures cluster into two categories:

\begin{itemize}[nosep]
    \item \textbf{Contradictory/adversarial premises} (18 cases, 53\%): Problems where the stated numbers are logically inconsistent, designed to confuse models.
    \item \textbf{Ambiguous referents} (16 cases, 47\%): Problems where pronouns or phrases like ``how many'' can refer to different quantities.
\end{itemize}

\paragraph{Case 1: Contradictory premise (score = 0.0).}

\begin{quote}
\small
\textbf{Problem}: ``A waiter had 3 customers. After some left he still had 4 customers. How many more customers stayed behind than those that left?''

\textbf{Expected}: 5.0 \quad \textbf{Predicted}: 2
\end{quote}

\noindent
\textit{Analysis.}
Starting with 3 and ``still having 4'' after some leave is impossible.
All five reasoning paths produced answers based on the literal (contradictory) text, and ScEnsemble converged on 2.
Even the Programmer verification step faithfully computed from the inconsistent numbers.
\textbf{Root cause}: SVAMP deliberately includes adversarially constructed problems. The workflow has no mechanism to flag or reinterpret contradictory premises.

\paragraph{Case 2: Distractor numbers (score = 0.0).}

\begin{quote}
\small
\textbf{Problem}: ``There were 61 parents and 177 pupils at the school concert. The program could seat 44 people. How many people were present?''

\textbf{Expected}: 238.0 \quad \textbf{Predicted}: 44
\end{quote}

\noindent
\textit{Analysis.}
The question asks ``how many people were present'' ($61 + 177 = 238$), but the model latched onto the distractor number (seating capacity = 44).
The ensemble reinforced this because all five paths were confused by the same irrelevant detail.
\textbf{Root cause}: SVAMP's adversarial construction places distractors that systematically mislead all paths simultaneously.

\subsection{AQuA: multiple-choice math reasoning}
\label{app:ood_aqua}
% ----------------------------------------------------------

\noindent
\textbf{Score}: 79.9\% (34 failures / 204 problems). \textbf{Gain over vanilla}: +27.0 points. \\
\textbf{Workflow}: 3$\times$Custom (solve MC) $\to$ ScEnsemble (vote) $\to$ Custom (extract letter).

AQuA shows the largest gain over vanilla, demonstrating that the ensemble+extraction pipeline is especially effective for multiple-choice math.
Failures split into reasoning errors (65\%) and option-mapping errors (35\%).

% --- AQuA failure type pie chart ---
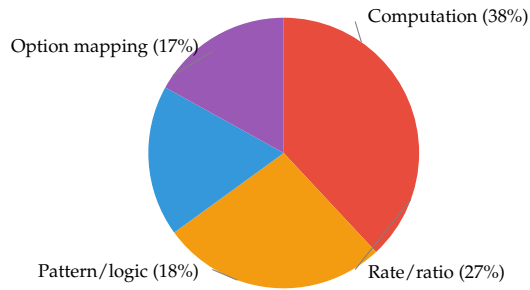
\begin{figure}[h]
\centering
\begin{tikzpicture}
\definecolor{pieA}{HTML}{E74C3C}
\definecolor{pieB}{HTML}{F39C12}
\definecolor{pieC}{HTML}{3498DB}
\definecolor{pieD}{HTML}{9B59B6}

% Pie chart
\def\radius{1.8}

\fill[pieA] (0,0) -- (90:\radius) arc (90:90-137:\radius) -- cycle;

\fill[pieB] (0,0) -- (90-137:\radius) arc (90-137:90-137-97:\radius) -- cycle;

\fill[pieC] (0,0) -- (90-234:\radius) arc (90-234:90-234-65:\radius) -- cycle;

\fill[pieD] (0,0) -- (90-299:\radius) arc (90-299:90-360:\radius) -- cycle;

% Labels with lines
\node[font=\scriptsize, anchor=west] at (1.0, 1.8) {Computation (38\%)};
\draw[gray, thin] (0.95, 1.75) -- (55:\radius);

\node[font=\scriptsize, anchor=west] at (1.0, -1.6) {Rate/ratio (27\%)};
\draw[gray, thin] (0.95, -1.55) -- (-20:\radius);

\node[font=\scriptsize, anchor=east] at (-1.0, -1.6) {Pattern/logic (18\%)};
\draw[gray, thin] (-0.95, -1.55) -- (-110:\radius);

\node[font=\scriptsize, anchor=east] at (-1.0, 1.4) {Option mapping (17\%)};
\draw[gray, thin] (-0.95, 1.35) -- (145:\radius);

\end{tikzpicture}
\caption{AQuA failure type distribution (34 failures). Computation errors dominate (38\%), followed by rate/ratio reasoning (27\%). Option mapping errors (17\%) occur when the model computes the correct numerical answer but selects the wrong multiple-choice letter.}
\label{fig:aqua_pie}
\end{figure}

\paragraph{Case 1: Compound vs.\ simple interest (score = 0.0).}

\begin{quote}
\small
\textbf{Problem}: ``If the population of a city increases by 5\% annually, what will be the population in 2 years if its current population is 78000? A)~81900 B)~85995 C)~85800 D)~90000 E)~None''

\textbf{Expected}: B \quad \textbf{Predicted}: C
\end{quote}

\noindent
\textit{Analysis.}
Correct: $78{,}000 \times 1.05^2 = 85{,}995$ (B).
Predicted: $78{,}000 \times 1.10 = 85{,}800$ (C, simple interest).
All three paths used simple rather than compound growth.
\textbf{Root cause}: Systematic reasoning bias---``increases by 5\% annually'' is treated as simple rather than compound growth.

\paragraph{Case 2: Non-arithmetic pattern puzzle (score = 0.0).}

\begin{quote}
\small
\textbf{Problem}: ``At my fruit stand, an orange costs 18, pineapple costs 27, grape costs 15. How much does a mango cost? A)~22 B)~15 C)~20 D)~18 E)~10''

\textbf{Expected}: B (15) \quad \textbf{Predicted}: C (20)
\end{quote}

\noindent
\textit{Analysis.}
Pattern: price = 3 $\times$ letter count (orange=6$\times$3=18, pineapple=9$\times$3=27, mango=5$\times$3=15).
The workflow's math-focused instructions (transferred from GSM8K/MATH demos) prime the model to seek arithmetic relationships between numbers, completely missing the letter-counting pattern.
\textbf{Root cause}: Strategy transfer mismatch---the transferred reasoning approach (numerical computation) is misaligned with non-arithmetic puzzles.

\paragraph{Case 3: Rate problem --- water lily doubling (score = 0.0).}

\begin{quote}
\small
\textbf{Problem}: ``A pond has water lilies that double every day. If the pond is completely covered on day 48, on what day was it half covered? A)~24 B)~36 C)~40 D)~47 E)~46''

\textbf{Expected}: D (47) \quad \textbf{Predicted}: E (46)
\end{quote}

\noindent
\textit{Analysis.}
If lilies double daily, half coverage was exactly one day before full coverage: day $48 - 1 = 47$ (D).
The model answered 46 (two days before), perhaps applying doubling twice.
\textbf{Root cause}: Exponential reasoning is unintuitive; the ensemble cannot correct this when all paths share the same misunderstanding.

% ----------------------------------------------------------
\subsection{AIME 2024--2025: competition mathematics}
\label{app:ood_aime}
% ----------------------------------------------------------

\noindent
\textbf{Score}: 14.6\% (41 failures / 48 problems). \textbf{Gain over vanilla}: +6.3 points. \\
\textbf{Workflow}: Programmer (code solve) $\to$ Custom (interpret) $\to$ 3$\times$Custom (alternatives) $\to$ ScEnsemble $\to$ Custom (extract \texttt{\textbackslash boxed\{\}}).

AIME is the hardest OOD benchmark, with an 85.4\% failure rate.
Even the most elaborate synthesized workflow (code+4-path ensemble+extraction) cannot compensate for GPT-4o-mini's limited mathematical reasoning on competition-level problems.
The +6.3 gain over vanilla shows that the workflow provides \emph{some} benefit (primarily from the code-verification step), but the ceiling is low.

% --- AIME: correct vs wrong magnitude comparison ---
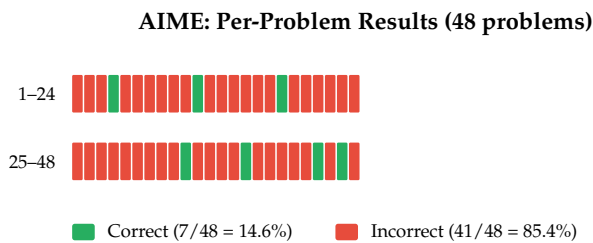
\begin{figure}[h]
\centering
\begin{tikzpicture}
\definecolor{hit}{HTML}{27AE60}
\definecolor{miss}{HTML}{E74C3C}

\def\blockw{0.16}
\def\blockh{0.5}

\node[anchor=south, font=\small\bfseries] at (3.9, 3.3) {AIME: Per-Problem Results (48 problems)};

% Row 1: problems 1-24
\foreach \i/\c in {
1/miss,2/miss,3/miss,4/hit,5/miss,6/miss,7/miss,8/miss,9/miss,10/miss,
11/hit,12/miss,13/miss,14/miss,15/miss,16/miss,17/miss,18/hit,19/miss,20/miss,
21/miss,22/miss,23/miss,24/miss} {
    \fill[\c, rounded corners=0.5pt] ({(\i-1)*\blockw}, 2.4) rectangle ({(\i-1)*\blockw+\blockw-0.02}, 2.4+\blockh);
}

% Row 2: problems 25-48
\foreach \i/\c in {
25/miss,26/miss,27/miss,28/miss,29/miss,30/miss,31/miss,32/miss,33/miss,34/hit,
35/miss,36/miss,37/miss,38/miss,39/hit,40/miss,41/miss,42/miss,43/miss,44/miss,
45/hit,46/miss,47/hit,48/miss} {
    \fill[\c, rounded corners=0.5pt] ({(\i-25)*\blockw}, 1.5) rectangle ({(\i-25)*\blockw+\blockw-0.02}, 1.5+\blockh);
}

% Labels
\node[anchor=east, font=\scriptsize] at (-0.1, 2.65) {1--24};
\node[anchor=east, font=\scriptsize] at (-0.1, 1.75) {25--48};

% Legend
\fill[hit, rounded corners=1pt] (0, 0.7) rectangle (0.3, 0.92);
\node[anchor=west, font=\scriptsize] at (0.35, 0.81) {Correct (7/48 = 14.6\%)};
\fill[miss, rounded corners=1pt] (3.5, 0.7) rectangle (3.8, 0.92);
\node[anchor=west, font=\scriptsize] at (3.85, 0.81) {Incorrect (41/48 = 85.4\%)};

\end{tikzpicture}
\caption{AIME per-problem results. Each cell represents one problem; green = correct, red = incorrect. Only 7 of 48 problems are solved correctly (scattered across problem indices), confirming that failures are not clustered in ``hard'' problem positions but reflect fundamental capability limits across diverse mathematical topics.}
\label{fig:aime_heatmap}
\end{figure}

\paragraph{Case 1: Combinatorial geometry (score = 0.0).}

\begin{quote}
\small
\textbf{Problem}: ``Find the number of rectangles that can be formed inside a fixed regular dodecagon where each side lies on a side or diagonal.''

\textbf{Expected}: 315 \quad \textbf{Predicted}: $\backslash$\texttt{boxed\{36\}}
\end{quote}

\noindent
\textit{Analysis.}
The Programmer operator generated code enumerating only axis-aligned rectangles (36), missing the rotated ones.
The three Custom paths also produced underestimates.
\textbf{Root cause}: Competition-level geometry exceeds GPT-4o-mini's capability, and the code-verification step implements an incomplete enumeration.

\paragraph{Case 2: Number theory (score = 0.0).}

\begin{quote}
\small
\textbf{Problem}: ``Let $A$ be the set of positive integer divisors of 2025. Let $B$ be a randomly selected subset of $A$. The probability that $B$ is a nonempty set with LCM equal to 2025 is $m/n$. Find $m+n$.''

\textbf{Expected}: 237 \quad \textbf{Predicted}: $\backslash$\texttt{boxed\{32777\}}
\end{quote}

\noindent
\textit{Analysis.}
The predicted answer is two orders of magnitude off, indicating a fundamental misapplication of inclusion-exclusion.
\textbf{Root cause}: Advanced number theory exceeds the execution model's reasoning depth.

\paragraph{Case 3: Walking speed optimization (score = 0.0).}

\begin{quote}
\small
\textbf{Problem}: ``Aya walks at different speeds during a 9-block walk. Blocks 1--3 at $s$ ft/s, blocks 4--6 at $2s$ ft/s, blocks 7--9 at a specified pattern. She takes 25\% more time than walking all 9 blocks at $s$ ft/s. Find $s^2$.''

\textbf{Expected}: 204 \quad \textbf{Predicted}: $\backslash$\texttt{boxed\{180\}}
\end{quote}

\noindent
\textit{Analysis.}
The Programmer code solved for $s^2$ but set up the time equation incorrectly, computing $180$ instead of $204$.
\textbf{Root cause}: Multi-step algebraic word problems with fraction arithmetic compound errors through the pipeline.

% ----------------------------------------------------------
\subsection{BigCodeBench: complex code generation}
\label{app:ood_bcb}
% ----------------------------------------------------------

\noindent
\textbf{Score}: 34.3\% (542 failures / 912 problems). \textbf{Gain over vanilla}: +2.7 points. \\
\textbf{Workflow}: 3$\times$CustomCodeGenerate $\to$ ScEnsemble $\to$ Test $\to$ if fail: retry.

BigCodeBench has the highest failure rate (59.4\%) but the story is nuanced: \textbf{71\% of failures are caused by missing Python modules}, not by incorrect code generation.
The true logic-error failure rate (excluding environment issues) is approximately 17\%.

% --- BigCodeBench error type breakdown ---
\begin{figure}[h]
\centering
\begin{tikzpicture}
\definecolor{envA}{HTML}{3498DB}   % pyparsing
\definecolor{envB}{HTML}{5DADE2}   % other module
\definecolor{logicA}{HTML}{E74C3C} % assertion
\definecolor{logicB}{HTML}{F1948A} % other logic
\definecolor{other}{HTML}{95A5A6}  % other

% Horizontal stacked bar
\def\bw{0.8}
\def\ml{10.0}

\fill[envA, rounded corners=2pt]  (0, 0)               rectangle ({\ml*0.535}, \bw);
\fill[envB, rounded corners=0pt]  ({\ml*0.535}, 0)      rectangle ({\ml*0.710}, \bw);
\fill[logicA, rounded corners=0pt]({\ml*0.710}, 0)      rectangle ({\ml*0.900}, \bw);
\fill[logicB, rounded corners=0pt]({\ml*0.900}, 0)      rectangle ({\ml*0.987}, \bw);
\fill[other, rounded corners=2pt] ({\ml*0.987}, 0)      rectangle ({\ml*1.0}, \bw);

% Labels inside bars
\node[font=\scriptsize, white] at ({\ml*0.535/2}, \bw/2) {\texttt{pyparsing} (290)};
\node[font=\scriptsize, white] at ({\ml*(0.535+0.710)/2}, \bw/2) {other (95)};
\node[font=\scriptsize, white] at ({\ml*(0.710+0.900)/2}, \bw/2) {assert (103)};
\node[font=\scriptsize, white] at ({\ml*(0.900+0.987)/2}, \bw/2) {47};

% Category brackets below
\draw[decorate, decoration={brace, amplitude=4pt, mirror}, thick, blue!70] (0, -0.15) -- ({\ml*0.710}, -0.15);
\node[font=\scriptsize, blue!70] at ({\ml*0.355}, -0.55) {\textbf{Environment issues (385 = 71\%)}};

\draw[decorate, decoration={brace, amplitude=4pt, mirror}, thick, red!70] ({\ml*0.710}, -0.15) -- ({\ml*0.987}, -0.15);
\node[font=\scriptsize, red!70] at ({\ml*0.849}, -0.55) {\textbf{Logic errors (150 = 28\%)}};

\end{tikzpicture}
\caption{BigCodeBench failure type breakdown (542 failures). The dominant failure mode is missing Python modules (71\%), particularly \texttt{pyparsing} (a transitive dependency of \texttt{matplotlib}). Excluding environment issues, the true logic-error failure rate drops from 59.4\% to approximately 17\%.}
\label{fig:bcb_breakdown}
\end{figure}
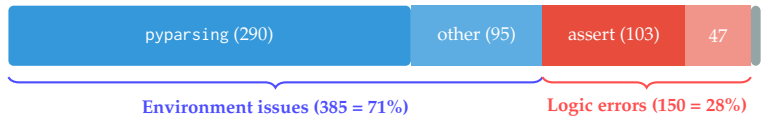

\paragraph{Case 1: Environment mismatch (score = 0.0; representative of 385 cases).}

\begin{quote}
\small
\textbf{Problem}: \texttt{task\_func(data)}: ``Generate a bar plot showing the frequency of letters.'' (requires \texttt{matplotlib})

\textbf{Predicted}: Logically correct code using \texttt{matplotlib.pyplot}

\textbf{Error}: \texttt{No module named `pyparsing'}
\end{quote}

\noindent
\textit{Analysis.}
The code is correct but the execution environment lacks \texttt{pyparsing} (a transitive dependency of \texttt{matplotlib}).
This accounts for 290 of the 542 failures.
\textbf{Root cause}: Infrastructure issue, not a model or workflow limitation.

\paragraph{Case 2: Logic bug --- encoding error (score = 0.0).}

\begin{quote}
\small
\textbf{Problem}: \texttt{task\_func(data)}: ``Apply StandardScaler normalization and return base64-encoded ASCII string.''

\textbf{Predicted}: Code that serializes raw float bytes before base64 encoding

\textbf{Error}: \texttt{UnicodeDecodeError: `ascii' codec can't decode byte 0xf0}
\end{quote}

\noindent
\textit{Analysis.}
The generated code base64-encodes raw numpy array bytes instead of first converting to a string representation.
The Test operator caught the failure, but the retry regenerated the same approach.
\textbf{Root cause}: Subtle serialization semantics that the test-driven retry cannot diagnose from the error message alone.

\subsection{Cross-dataset failure patterns}
\label{app:ood_patterns}

Table~\ref{tab:ood_failure_summary} summarizes the failure landscape across all four OOD benchmarks.

\begin{table}[h]
\centering
% \small
% \large
\renewcommand{\arraystretch}{1.15}
\begin{tabular}{@{} l r r >{\raggedright\arraybackslash}p{3.0cm} >{\raggedright\arraybackslash}p{4.5cm} @{}}
\toprule
\textbf{Dataset} & \textbf{Fail Rate}  & \textbf{Primary Mode} & \textbf{Why workflow helps / doesn't help} \\
\midrule
SVAMP         & 4.2\%   & Adversarial wording     & Ensemble catches many errors; adversarial premises defeat all paths \\
AQuA          & 16.7\%  & Computation error       & Ensemble + letter extraction very effective; compound reasoning still fails \\
AIME          & 85.4\%   & Capability ceiling      & Code+ensemble adds marginal value; problem difficulty exceeds model capability \\
BigCodeBench  & 59.4\%  & Environment (71\%)      & Workflow can't fix missing modules; logic errors partially caught by test+retry \\
\bottomrule
\end{tabular}
\captionsetup{font=large}
\caption{OOD failure summary. The workflow's effectiveness correlates with the gap between problem difficulty and execution model capability: large gains where the model \emph{can} solve problems with better structure (SVAMP, AQuA), diminishing returns where the model fundamentally cannot (AIME), and near-zero gains where failures are environmental (BigCodeBench).}
\label{tab:ood_failure_summary}
\end{table}

Three cross-cutting patterns emerge:

\begin{enumerate}
    \item \textbf{Transfer effectiveness scales with task similarity.} SVAMP and AQuA (math benchmarks similar to GSM8K/MATH) show the largest gains (+14.8, +27.0). BigCodeBench (complex code tasks dissimilar from HumanEval/MBPP) shows the smallest (+2.7). The synthesized workflow's strategies transfer best when the OOD task shares structural similarities with the demonstration tasks.

    \item \textbf{Ensemble voting has diminishing returns on hard problems.} On SVAMP (easy), ensemble corrects many individual reasoning errors, yielding 95.8\% accuracy. On AIME (hard), all ensemble paths fail on the same problems, providing only +6.3 over vanilla. Self-consistency voting is effective when errors are \emph{random} (different paths make different mistakes) but not when errors are \emph{systematic} (all paths share the same capability limitation).

    \item \textbf{Infrastructure failures mask model capability.} BigCodeBench's 59.4\% failure rate suggests poor transfer, but 71\% of failures are environmental (\texttt{pyparsing}, etc.). Adjusting for this, the true logic-error rate is $\sim$17\%, comparable to MBPP's 17.9\%. The workflow's code generation quality transfers well; the bottleneck is the execution sandbox, not the synthesis pipeline.
\end{enumerate}

\section{Dataset statistics and representative examples}
\label{app:dataset_stats}
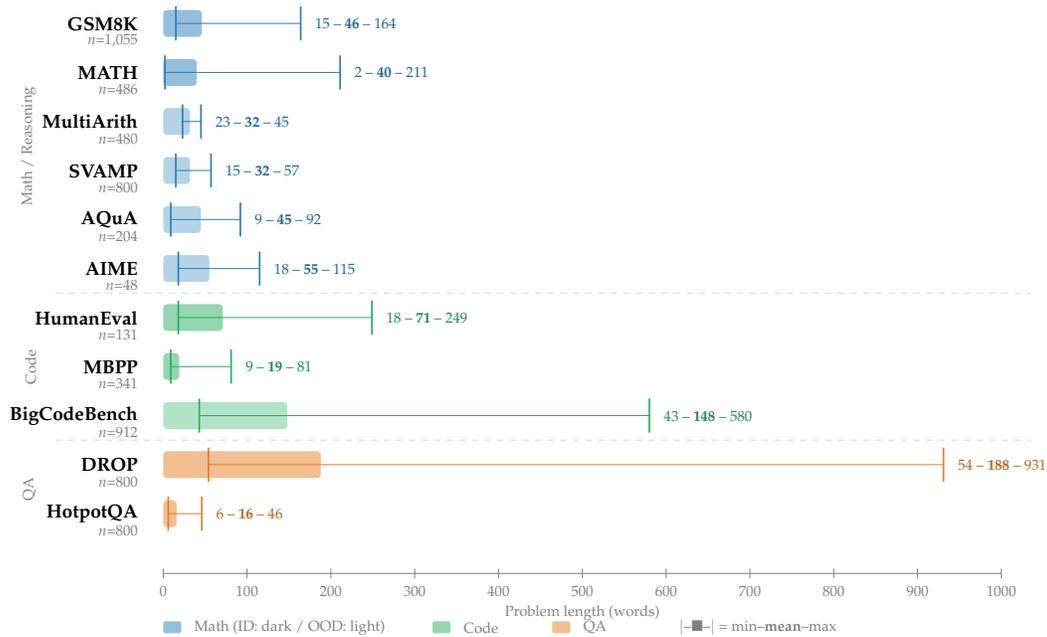
\begin{figure}[p]
\centering
\resizebox{\textwidth}{!}{%
\begin{tikzpicture}[xscale=1, yscale=1]
\definecolor{mathc}{HTML}{2980B9}
\definecolor{codec}{HTML}{27AE60}
\definecolor{qac}{HTML}{E67E22}

\def\scl{0.014}
\def\bh{0.45}       % bar height
\def\gap{0.82}      % vertical spacing between rows

% --- GSM8K: mean=46, min=15, max=164, n=1055 ---
\node[font=\small\bfseries, anchor=east] at (-0.3, 10*\gap) {GSM8K};
\node[font=\scriptsize, gray, anchor=east] at (-0.3, 10*\gap-0.28) {\textit{n}=1{,}055};
\fill[mathc!50, rounded corners=2pt] (0, 10*\gap-\bh/2) rectangle ({46*\scl}, 10*\gap+\bh/2);
\draw[mathc, thick] ({15*\scl}, 10*\gap-\bh/2-0.06) -- ({15*\scl}, 10*\gap+\bh/2+0.06);
\draw[mathc, thick] ({164*\scl}, 10*\gap-\bh/2-0.06) -- ({164*\scl}, 10*\gap+\bh/2+0.06);
\draw[mathc] ({15*\scl}, 10*\gap) -- ({46*\scl}, 10*\gap);
\draw[mathc] ({46*\scl}, 10*\gap) -- ({164*\scl}, 10*\gap);
\node[font=\scriptsize, mathc!80!black, anchor=west] at ({164*\scl+0.12}, 10*\gap) {15 -- \textbf{46} -- 164};

% --- MATH: mean=40, min=2, max=211, n=486 ---
\node[font=\small\bfseries, anchor=east] at (-0.3, 9*\gap) {MATH};
\node[font=\scriptsize, gray, anchor=east] at (-0.3, 9*\gap-0.28) {\textit{n}=486};
\fill[mathc!50, rounded corners=2pt] (0, 9*\gap-\bh/2) rectangle ({40*\scl}, 9*\gap+\bh/2);
\draw[mathc, thick] ({2*\scl}, 9*\gap-\bh/2-0.06) -- ({2*\scl}, 9*\gap+\bh/2+0.06);
\draw[mathc, thick] ({211*\scl}, 9*\gap-\bh/2-0.06) -- ({211*\scl}, 9*\gap+\bh/2+0.06);
\draw[mathc] ({2*\scl}, 9*\gap) -- ({40*\scl}, 9*\gap);
\draw[mathc] ({40*\scl}, 9*\gap) -- ({211*\scl}, 9*\gap);
\node[font=\scriptsize, mathc!80!black, anchor=west] at ({211*\scl+0.12}, 9*\gap) {2 -- \textbf{40} -- 211};

% --- MultiArith: mean=32, min=23, max=45, n=480 ---
\node[font=\small\bfseries, anchor=east] at (-0.3, 8*\gap) {MultiArith};
\node[font=\scriptsize, gray, anchor=east] at (-0.3, 8*\gap-0.28) {\textit{n}=480};
\fill[mathc!35, rounded corners=2pt] (0, 8*\gap-\bh/2) rectangle ({32*\scl}, 8*\gap+\bh/2);
\draw[mathc, thick] ({23*\scl}, 8*\gap-\bh/2-0.06) -- ({23*\scl}, 8*\gap+\bh/2+0.06);
\draw[mathc, thick] ({45*\scl}, 8*\gap-\bh/2-0.06) -- ({45*\scl}, 8*\gap+\bh/2+0.06);
\draw[mathc] ({23*\scl}, 8*\gap) -- ({32*\scl}, 8*\gap);
\draw[mathc] ({32*\scl}, 8*\gap) -- ({45*\scl}, 8*\gap);
\node[font=\scriptsize, mathc!80!black, anchor=west] at ({45*\scl+0.12}, 8*\gap) {23 -- \textbf{32} -- 45};

% --- SVAMP: mean=32, min=15, max=57, n=800 ---
\node[font=\small\bfseries, anchor=east] at (-0.3, 7*\gap) {SVAMP};
\node[font=\scriptsize, gray, anchor=east] at (-0.3, 7*\gap-0.28) {\textit{n}=800};
\fill[mathc!35, rounded corners=2pt] (0, 7*\gap-\bh/2) rectangle ({32*\scl}, 7*\gap+\bh/2);
\draw[mathc, thick] ({15*\scl}, 7*\gap-\bh/2-0.06) -- ({15*\scl}, 7*\gap+\bh/2+0.06);
\draw[mathc, thick] ({57*\scl}, 7*\gap-\bh/2-0.06) -- ({57*\scl}, 7*\gap+\bh/2+0.06);
\draw[mathc] ({15*\scl}, 7*\gap) -- ({32*\scl}, 7*\gap);
\draw[mathc] ({32*\scl}, 7*\gap) -- ({57*\scl}, 7*\gap);
\node[font=\scriptsize, mathc!80!black, anchor=west] at ({57*\scl+0.12}, 7*\gap) {15 -- \textbf{32} -- 57};

% --- AQuA: mean=45, min=9, max=92, n=204 ---
\node[font=\small\bfseries, anchor=east] at (-0.3, 6*\gap) {AQuA};
\node[font=\scriptsize, gray, anchor=east] at (-0.3, 6*\gap-0.28) {\textit{n}=204};
\fill[mathc!35, rounded corners=2pt] (0, 6*\gap-\bh/2) rectangle ({45*\scl}, 6*\gap+\bh/2);
\draw[mathc, thick] ({9*\scl}, 6*\gap-\bh/2-0.06) -- ({9*\scl}, 6*\gap+\bh/2+0.06);
\draw[mathc, thick] ({92*\scl}, 6*\gap-\bh/2-0.06) -- ({92*\scl}, 6*\gap+\bh/2+0.06);
\draw[mathc] ({9*\scl}, 6*\gap) -- ({45*\scl}, 6*\gap);
\draw[mathc] ({45*\scl}, 6*\gap) -- ({92*\scl}, 6*\gap);
\node[font=\scriptsize, mathc!80!black, anchor=west] at ({92*\scl+0.12}, 6*\gap) {9 -- \textbf{45} -- 92};

% --- AIME: mean=55, min=18, max=115, n=48 ---
\node[font=\small\bfseries, anchor=east] at (-0.3, 5*\gap) {AIME};
\node[font=\scriptsize, gray, anchor=east] at (-0.3, 5*\gap-0.28) {\textit{n}=48};
\fill[mathc!35, rounded corners=2pt] (0, 5*\gap-\bh/2) rectangle ({55*\scl}, 5*\gap+\bh/2);
\draw[mathc, thick] ({18*\scl}, 5*\gap-\bh/2-0.06) -- ({18*\scl}, 5*\gap+\bh/2+0.06);
\draw[mathc, thick] ({115*\scl}, 5*\gap-\bh/2-0.06) -- ({115*\scl}, 5*\gap+\bh/2+0.06);
\draw[mathc] ({18*\scl}, 5*\gap) -- ({55*\scl}, 5*\gap);
\draw[mathc] ({55*\scl}, 5*\gap) -- ({115*\scl}, 5*\gap);
\node[font=\scriptsize, mathc!80!black, anchor=west] at ({115*\scl+0.12}, 5*\gap) {18 -- \textbf{55} -- 115};

% --- HumanEval: mean=71, min=18, max=249, n=131 ---
\node[font=\small\bfseries, anchor=east] at (-0.3, 4*\gap) {HumanEval};
\node[font=\scriptsize, gray, anchor=east] at (-0.3, 4*\gap-0.28) {\textit{n}=131};
\fill[codec!50, rounded corners=2pt] (0, 4*\gap-\bh/2) rectangle ({71*\scl}, 4*\gap+\bh/2);
\draw[codec, thick] ({18*\scl}, 4*\gap-\bh/2-0.06) -- ({18*\scl}, 4*\gap+\bh/2+0.06);
\draw[codec, thick] ({249*\scl}, 4*\gap-\bh/2-0.06) -- ({249*\scl}, 4*\gap+\bh/2+0.06);
\draw[codec] ({18*\scl}, 4*\gap) -- ({71*\scl}, 4*\gap);
\draw[codec] ({71*\scl}, 4*\gap) -- ({249*\scl}, 4*\gap);
\node[font=\scriptsize, codec!80!black, anchor=west] at ({249*\scl+0.12}, 4*\gap) {18 -- \textbf{71} -- 249};

% --- MBPP: mean=19, min=9, max=81, n=341 ---
\node[font=\small\bfseries, anchor=east] at (-0.3, 3*\gap) {MBPP};
\node[font=\scriptsize, gray, anchor=east] at (-0.3, 3*\gap-0.28) {\textit{n}=341};
\fill[codec!50, rounded corners=2pt] (0, 3*\gap-\bh/2) rectangle ({19*\scl}, 3*\gap+\bh/2);
\draw[codec, thick] ({9*\scl}, 3*\gap-\bh/2-0.06) -- ({9*\scl}, 3*\gap+\bh/2+0.06);
\draw[codec, thick] ({81*\scl}, 3*\gap-\bh/2-0.06) -- ({81*\scl}, 3*\gap+\bh/2+0.06);
\draw[codec] ({9*\scl}, 3*\gap) -- ({19*\scl}, 3*\gap);
\draw[codec] ({19*\scl}, 3*\gap) -- ({81*\scl}, 3*\gap);
\node[font=\scriptsize, codec!80!black, anchor=west] at ({81*\scl+0.12}, 3*\gap) {9 -- \textbf{19} -- 81};

% --- BigCodeBench: mean=148, min=43, max=580, n=912 ---
\node[font=\small\bfseries, anchor=east] at (-0.3, 2*\gap) {BigCodeBench};
\node[font=\scriptsize, gray, anchor=east] at (-0.3, 2*\gap-0.28) {\textit{n}=912};
\fill[codec!35, rounded corners=2pt] (0, 2*\gap-\bh/2) rectangle ({148*\scl}, 2*\gap+\bh/2);
\draw[codec, thick] ({43*\scl}, 2*\gap-\bh/2-0.06) -- ({43*\scl}, 2*\gap+\bh/2+0.06);
\draw[codec, thick] ({580*\scl}, 2*\gap-\bh/2-0.06) -- ({580*\scl}, 2*\gap+\bh/2+0.06);
\draw[codec] ({43*\scl}, 2*\gap) -- ({148*\scl}, 2*\gap);
\draw[codec] ({148*\scl}, 2*\gap) -- ({580*\scl}, 2*\gap);
\node[font=\scriptsize, codec!80!black, anchor=west] at ({580*\scl+0.12}, 2*\gap) {43 -- \textbf{148} -- 580};

\node[font=\small\bfseries, anchor=east] at (-0.3, 1*\gap) {DROP};
\node[font=\scriptsize, gray, anchor=east] at (-0.3, 1*\gap-0.28) {\textit{n}=800};
\fill[qac!50, rounded corners=2pt] (0, 1*\gap-\bh/2) rectangle ({188*\scl}, 1*\gap+\bh/2);
\draw[qac, thick] ({54*\scl}, 1*\gap-\bh/2-0.06) -- ({54*\scl}, 1*\gap+\bh/2+0.06);
\draw[qac, thick] ({931*\scl}, 1*\gap-\bh/2-0.06) -- ({931*\scl}, 1*\gap+\bh/2+0.06);
\draw[qac] ({54*\scl}, 1*\gap) -- ({188*\scl}, 1*\gap);
\draw[qac] ({188*\scl}, 1*\gap) -- ({931*\scl}, 1*\gap);
\node[font=\scriptsize, qac!80!black, anchor=west] at ({931*\scl+0.12}, 1*\gap) {54 -- \textbf{188} -- 931};

\node[font=\small\bfseries, anchor=east] at (-0.3, 0) {HotpotQA};
\node[font=\scriptsize, gray, anchor=east] at (-0.3, -0.28) {\textit{n}=800};
\fill[qac!50, rounded corners=2pt] (0, -\bh/2) rectangle ({16*\scl}, \bh/2);
\draw[qac, thick] ({6*\scl}, -\bh/2-0.06) -- ({6*\scl}, \bh/2+0.06);
\draw[qac, thick] ({46*\scl}, -\bh/2-0.06) -- ({46*\scl}, \bh/2+0.06);
\draw[qac] ({6*\scl}, 0) -- ({16*\scl}, 0);
\draw[qac] ({16*\scl}, 0) -- ({46*\scl}, 0);
\node[font=\scriptsize, qac!80!black, anchor=west] at ({46*\scl+0.12}, 0) {6 -- \textbf{16} -- 46};

\draw[gray!30, thin, dashed] (-1.8, {4.5*\gap}) -- (14.5, {4.5*\gap});
\draw[gray!30, thin, dashed] (-1.8, {1.5*\gap}) -- (14.5, {1.5*\gap});
\node[font=\scriptsize, gray, rotate=90, anchor=south] at (-2.0, {7.5*\gap}) {Math / Reasoning};
\node[font=\scriptsize, gray, rotate=90, anchor=south] at (-2.0, {3*\gap}) {Code};
\node[font=\scriptsize, gray, rotate=90, anchor=south] at (-2.0, {0.5*\gap}) {QA};

% ===== Scale reference axis =====
\draw[gray, thin] (0, -1.0) -- (14, -1.0);
\foreach \x in {0, 100, 200, 300, 400, 500, 600, 700, 800, 900, 1000} {
    \draw[gray] ({\x*\scl}, -0.92) -- ({\x*\scl}, -1.08);
    \node[font=\scriptsize, gray] at ({\x*\scl}, -1.3) {\x};
}
\node[font=\scriptsize, gray] at ({500*\scl}, -1.65) {Problem length (words)};

% ===== Legend =====
\fill[mathc!50, rounded corners=1pt] (0, -2.0) rectangle (0.3, -1.8);
\node[font=\scriptsize, gray, anchor=west] at (0.4, -1.9) {Math (ID: dark / OOD: light)};
\fill[codec!50, rounded corners=1pt] (4.5, -2.0) rectangle (4.8, -1.8);
\node[font=\scriptsize, gray, anchor=west] at (4.9, -1.9) {Code};
\fill[qac!50, rounded corners=1pt] (6.5, -2.0) rectangle (6.8, -1.8);
\node[font=\scriptsize, gray, anchor=west] at (6.9, -1.9) {QA};
\node[font=\scriptsize, gray, anchor=west] at (8.5, -1.9) {$\vert$--$\blacksquare$--$\vert$ = min--\textbf{mean}--max};

\end{tikzpicture}%
}
\caption{Problem length distribution (words) for all 11 benchmarks. Filled bars show the mean; whiskers extend to min and max. DROP has the longest problems (passages averaging 188 words, max 931). BigCodeBench is second (code docstrings averaging 148 words, max 580). MATH has the widest variance among math benchmarks (2--211 words), reflecting diverse problem types. HotpotQA and MBPP have the shortest prompts (mean 16 and 19 words, respectively). OOD benchmarks (lighter shading) have comparable length distributions to their ID counterparts, confirming that OOD difficulty comes from task structure, not input length.}
\label{fig:problem_lengths}
\end{figure}

This section supplements the benchmark overview in the main paper with problem-level statistics.
Table~\ref{tab:dataset_length} reports the problem length distribution for all benchmarks, and Figure~\ref{fig:problem_lengths} visualizes the full min--mean--max range.

\begin{table}[h]
\centering
\small
% \large
\renewcommand{\arraystretch}{1.2}
\begin{tabular}{l c r >{\raggedleft\arraybackslash}p{1cm} >{\raggedleft\arraybackslash}p{1cm} >{\raggedleft\arraybackslash}p{1cm}}
\toprule
\textbf{Benchmark} & \textbf{ID/OOD} & \textbf{Test} & \multicolumn{3}{c}{\textbf{Problem Length (words)}} \\
\cmidrule(lr){4-6}
 & & & \textbf{Mean} & \textbf{Min} & \textbf{Max} \\
\midrule
\rowcolor{blue!3}
GSM8K       & ID  & 1{,}055 & 46  & 15  & 164 \\
\rowcolor{blue!3}
MATH        & ID  & 486     & 40  & 2   & 211 \\
\rowcolor{green!3}
HumanEval   & ID  & 131     & 71  & 18  & 249 \\
\rowcolor{green!3}
MBPP        & ID  & 341     & 19  & 9   & 81 \\
\rowcolor{orange!3}
DROP        & ID  & 800     & 188 & 54  & 931 \\
\rowcolor{orange!3}
HotpotQA    & ID  & 800     & 16  & 6   & 46 \\
\midrule
\rowcolor{blue!3}
MultiArith  & OOD & 480     & 32  & 23  & 45 \\
\rowcolor{blue!3}
SVAMP       & OOD & 800     & 32  & 15  & 57 \\
\rowcolor{blue!3}
AQuA        & OOD & 204     & 45  & 9   & 92 \\
\rowcolor{blue!3}
AIME        & OOD & 48      & 55  & 18  & 115 \\
\rowcolor{green!3}
BigCodeBench & OOD & 912    & 148 & 43  & 580 \\
\bottomrule
\end{tabular}
% \captionsetup{font=large}
\caption{Problem length statistics for all 11 benchmarks. DROP has the longest inputs (passages averaging 188 words). Math benchmarks have shorter, more uniform inputs. OOD benchmarks have comparable length distributions to their ID counterparts, confirming that OOD difficulty comes from task structure, not input length.}
\label{tab:dataset_length}
\end{table}

\subsection{MATH benchmark: topic distribution}

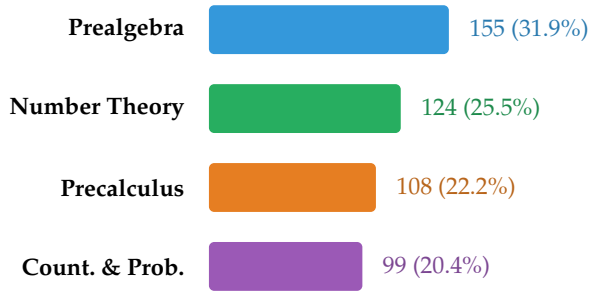
\begin{figure}[H]
\centering
\begin{tikzpicture}
\definecolor{pre}{HTML}{3498DB}
\definecolor{nt}{HTML}{27AE60}
\definecolor{pc}{HTML}{E67E22}
\definecolor{cp}{HTML}{9B59B6}

\def\bw{0.65}
\def\ml{10.0}
\def\rg{1.05}  % row gap

\node[font=\small\bfseries, anchor=east] at (-0.2, 3*\rg+\bw/2) {Prealgebra};
\fill[pre, rounded corners=2pt] (0, 3*\rg) rectangle ({\ml*0.319}, 3*\rg+\bw);
\node[font=\small, pre!80!black, anchor=west] at ({\ml*0.319+0.15}, 3*\rg+\bw/2) {155 (31.9\%)};

\node[font=\small\bfseries, anchor=east] at (-0.2, 2*\rg+\bw/2) {Number Theory};
\fill[nt, rounded corners=2pt] (0, 2*\rg) rectangle ({\ml*0.255}, 2*\rg+\bw);
\node[font=\small, nt!80!black, anchor=west] at ({\ml*0.255+0.15}, 2*\rg+\bw/2) {124 (25.5\%)};

\node[font=\small\bfseries, anchor=east] at (-0.2, 1*\rg+\bw/2) {Precalculus};
\fill[pc, rounded corners=2pt] (0, 1*\rg) rectangle ({\ml*0.222}, 1*\rg+\bw);
\node[font=\small, pc!80!black, anchor=west] at ({\ml*0.222+0.15}, 1*\rg+\bw/2) {108 (22.2\%)};

\node[font=\small\bfseries, anchor=east] at (-0.2, 0*\rg+\bw/2) {Count.\ \& Prob.};
\fill[cp, rounded corners=2pt] (0, 0*\rg) rectangle ({\ml*0.204}, 0*\rg+\bw);
\node[font=\small, cp!80!black, anchor=west] at ({\ml*0.204+0.15}, 0*\rg+\bw/2) {99 (20.4\%)};

\end{tikzpicture}
\captionsetup{font=large}
\caption{Topic distribution in the MATH test set (486 Level-5 problems). All four categories are roughly balanced, with Prealgebra slightly overrepresented.}
\label{fig:math_topics}
\end{figure}

The MATH test set (Level 5, highest difficulty) spans four topic areas (Figure~\ref{fig:math_topics}).

\section{Reproducibility details}
\label{app:reproducibility}

\subsection{Operator library}

\method{} reuses the operator library introduced by AFlow~\citep{zhang2025aflow}.
Table~\ref{tab:operators} defines the six operators available to the workflow synthesizer.

\begin{table}[ht]
\centering
% \small
\large
\renewcommand{\arraystretch}{1.15}
\resizebox{1\textwidth}{!}{%
\begin{tabular}{@{} l >{\raggedright}p{3.5cm} >{\raggedright}p{3.5cm} >{\raggedright\arraybackslash}p{3.5cm} @{}}
\toprule
\textbf{Operator} & \textbf{Input} & \textbf{Output} & \textbf{Description} \\
\midrule
Custom & text + instruction & text response & General-purpose text generation with custom instruction \\
AnswerGenerate & question text & thought + answer & Structured reasoning with thought chain \\
Programmer & problem description & code + output & Generates and executes Python code \\
CustomCodeGenerate & problem + entry point & code string & Generates code matching a function signature \\
ScEnsemble & list of solutions & best solution & Majority voting / selection \\
Test & code + test cases & pass/fail + output & Executes code against test cases \\
\bottomrule
\end{tabular}
}
\captionsetup{font=large}
\caption{Operator library specification. Each operator is a reusable building block with a fixed interface. The workflow synthesizer composes these operators into task-specific pipelines.}
\label{tab:operators}
\end{table}

\subsection{Computational cost}

\label{app:cost}
\begin{table}[H]
\centering
% \large
\small
\renewcommand{\arraystretch}{1.8}
\setlength{\tabcolsep}{10pt}
\captionsetup{font=large}
\caption{Computational cost breakdown per benchmark during execution. All costs are for GPT-4o-mini API calls. The total experiment cost across all 9 benchmarks is approximately \$6.08. Workflow synthesis cost (one-time, per seed) is negligible ($<$\$0.01 per benchmark).}
\begin{tabular*}{\textwidth}{@{\extracolsep{\fill}} l r r r r @{}}
\toprule
\textbf{Benchmark} & \textbf{Problems} & \textbf{LLM Calls} & \textbf{Avg Cost} & \textbf{Total Cost} \\
 & & \textbf{per Problem} & \textbf{per Problem} & \\
\midrule
\multicolumn{5}{@{}l}{\large\textbf{In-Distribution}} \\[4pt]
GSM8K       & 1{,}055 & 3     & \$0.0011 & \$1.20 \\
MATH        & 486     & 7     & \$0.0036 & \$1.73 \\
HumanEval   & 131     & 2--3  & \$0.0002 & \$0.03 \\
MBPP        & 341     & 3--4  & \$0.0001 & \$0.04 \\[4pt]
\midrule
\multicolumn{5}{@{}l}{\large\textbf{Out-of-Distribution}} \\[4pt]
MultiArith  & 480     & 2     & \$0.0002 & \$0.08 \\
SVAMP       & 800     & 7     & \$0.0010 & \$0.80 \\
AQuA        & 204     & 5     & \$0.0012 & \$0.24 \\
AIME        & 48      & 7     & \$0.0030 & \$0.14 \\
BigCodeBench & 912    & 5--6  & \$0.0020 & \$1.82 \\[4pt]
\midrule
\multicolumn{4}{@{}l}{\textbf{Total (all benchmarks)}} & \textbf{\$6.08} \\
\bottomrule
\end{tabular*}

\vspace{14pt}

\label{tab:costs}
\end{table}

\begin{table}[htbp]
\centering
\large
\renewcommand{\arraystretch}{1.2}
\setlength{\tabcolsep}{4.5pt}
\captionsetup{font=large}
\caption{Detailed cost and time breakdown for the four main benchmarks. \textbf{Optimization} includes one-time distillation and synthesis; \textbf{Execution} is running the synthesized workflow on the full test set. All costs are GPT-4o-mini API calls. Optimization accounts for less than $0.6\%$ of total expenditure and completes in under 11 seconds per benchmark.}
\label{tab:cost_time_breakdown}
\begin{tabular}{l ccc c cc c c}
\toprule
& \multicolumn{3}{c}{\textbf{Optimization (one-time)}} && \multicolumn{2}{c}{\textbf{Execution}} && \\
\cmidrule(lr){2-4} \cmidrule(lr){6-7}
\textbf{Benchmark} & Distill (\$) & Synth (\$) & Time && Cost (\$) & Time && \textbf{Total (\$)} \\
\midrule
GSM8K     & 0.001 & 0.003 & 10.6\,s && 1.199 & 35.1\,min && 1.203 \\
MATH      & 0.001 & 0.003 & \phantom{0}8.1\,s && 1.729 & \phantom{0}6.8\,min && 1.733 \\
HumanEval & 0.001 & 0.003 & \phantom{0}7.8\,s && 0.032 & \phantom{0}1.5\,min && 0.036 \\
MBPP      & 0.001 & 0.003 & \phantom{0}7.0\,s && 0.044 & \phantom{0}1.8\,min && 0.048 \\
\midrule
\textbf{Total} & \textbf{0.005} & \textbf{0.012} & \textbf{33.5\,s} && \textbf{3.004} & \textbf{45.2\,min} && \textbf{3.021} \\
\bottomrule
\end{tabular}
\end{table}

\section{Cost Accounting and Amortized Scaling}
\label{app:cost_accounting}
 
We provide a formal accounting of the cost structure underlying SWIFT, addressing whether the efficiency comparison in Table~\ref{tab:efficiency_compact} should also include the upstream search trajectories consumed during offline distillation.
 
\paragraph{Source-task trajectories as sunk costs.}
Under the strict leave-one-out protocol, the trajectories available to SWIFT when synthesizing a workflow for a target task $\mathcal{T}_{\text{target}}$ are produced exclusively from the optimization of other source tasks $\mathcal{T}_{\text{src}} = \mathcal{T}_{\text{all}} \setminus \{\mathcal{T}_{\text{target}}\}$.
For example, when the target is GSM8K, the trajectory pool is constructed from optimization runs on MATH, HumanEval, MBPP, DROP, and HotpotQA.
These search trajectories are byproducts of solving their respective source tasks: AFlow spends \$22.50 on MATH in order to optimize \emph{MATH's own workflow}, not to serve SWIFT.
Their cost is therefore already justified by source-task utility, independent of whether SWIFT exists.
 
\paragraph{Scaling law comparison.}
Let $C_{\text{search}}(T)$ denote the optimization cost of a search-based method on task $T$, and let $C_{\text{synth}}$ denote SWIFT's per-task synthesis cost after offline distillation.
For $n$ target tasks, the two paradigms incur:
\begin{align}
    C^{\text{search}}_{\text{total}} &= \sum_{i=1}^{n} C_{\text{search}}(T_i) \;\approx\; n \cdot C_{\text{search}}, \label{eq:search-cost} \\[4pt]
    C^{\text{amortized}}_{\text{total}} &= \underbrace{C_{\text{source}}}_{\text{one-time}} + \; n \cdot C_{\text{synth}}, \label{eq:amortized-cost}
\end{align}
where $C_{\text{source}} = \sum_{T \in \mathcal{T}_{\text{src}}} C_{\text{search}}(T)$ is the total cost of collecting source-task trajectories.
In our setting, $C_{\text{synth}} \approx \$0.004$, four orders of magnitude smaller than $C_{\text{search}} \approx \$22.50$.
 
\paragraph{Break-even analysis.}
Even under a conservative accounting that charges $C_{\text{source}}$ entirely to SWIFT, amortized synthesis becomes cheaper when
\begin{equation}
    C_{\text{source}} + n \cdot C_{\text{synth}} \;<\; n \cdot C_{\text{search}},
\end{equation}
yielding a break-even point of
\begin{equation}
    n^{*} = \frac{C_{\text{source}}}{C_{\text{search}} - C_{\text{synth}}} \;\approx\; \frac{112.50}{22.50 - 0.004} \;\approx\; 5.
\end{equation}
That is, even when the full source-task budget is attributed to SWIFT, amortized synthesis is strictly cheaper after five downstream tasks.
This is a conservative estimate: under the LOO interpretation---where source-task search costs are sunk---the break-even is immediate ($n^{*} = 1$).
In practice, our evaluation already spans 9 benchmarks (5 ID + 4 OOD), well beyond the conservative threshold.
In realistic deployment, where tens or hundreds of new tasks arise over time, the marginal cost of each additional workflow (\$0.004) renders the design step effectively free.
 
\paragraph{From cost reduction to structural insight.}
The cost advantage of amortized synthesis is not merely an engineering convenience; it reflects a structural property of the workflow design landscape.
Per-task search discovers that math tasks converge to multi-path ensembles and code tasks converge to test-driven retry loops, then discards this cross-task regularity when the next task arrives, paying the full search cost again.
SWIFT makes this structural knowledge persistent and transferable.
From this perspective, the leave-one-out protocol serves a dual purpose: it is both an anti-leakage safeguard \emph{and} a direct empirical test of whether optimization knowledge discovered on one set of tasks can reduce the cost of designing solvers for another.
The affirmative answer, higher accuracy at negligible marginal cost, confirms that per-task search is not only expensive but structurally wasteful.
 
Table~\ref{tab:cost_time_breakdown} provides a detailed per-benchmark breakdown of distillation, synthesis, and execution costs.

\section{Effect of Output Contracts on Synthesized Workflow Topology}
\label{app:contract_topology}

\begin{figure}[htbp]
\centering
\begin{tikzpicture}[
    opnode/.style={rectangle, draw, rounded corners=3pt, minimum width=1.6cm, minimum height=0.5cm, font=\tiny, align=center},
    opbest/.style={opnode, fill=green!15, draw=green!60!black},
    opworse/.style={opnode, fill=red!10, draw=red!60!black},
    arr/.style={->, >=stealth, semithick},
    lbl/.style={font=\scriptsize\bfseries, align=center},
]

% ==========================================
% GSM8K WITH Contract (left column)
% ==========================================
\node[lbl] at (1.5, 0) (g_tw) {\textsc{GSM8K} — \textit{With $\mathcal{C}$} (Acc\,=\,94.3)};
\node[opbest] at (1.5, -0.8) (g_ag) {AnswerGenerate};
\node[opbest] at (1.5, -1.8) (g_cu) {Custom (extract)};
\draw[arr] (g_ag) -- (g_cu);

% ==========================================
% GSM8K WITHOUT Contract (right column)
% ==========================================
\node[lbl] at (7.5, 0) (g_two) {\textsc{GSM8K} — \textit{w/o $\mathcal{C}$} (Acc\,=\,89.9)};
\node[opworse] at (5.5, -0.8) (g_a0) {AnswerGen};
\node[opworse] at (7.5, -0.8) (g_a1) {AnswerGen};
\node[opworse] at (9.5, -0.8) (g_a2) {AnswerGen};
\node[opworse] at (7.5, -1.8) (g_en) {ScEnsemble};
\node[opworse] at (7.5, -2.8) (g_c2) {Custom (extract)};
\draw[arr] (g_a0) -- (g_en);
\draw[arr] (g_a1) -- (g_en);
\draw[arr] (g_a2) -- (g_en);
\draw[arr] (g_en) -- (g_c2);

% ==========================================
% HumanEval WITH Contract (left column)
% ==========================================
\node[lbl] at (1.5, -4.0) (h_tw) {\textsc{HumanEval} — \textit{With $\mathcal{C}$} (Acc\,=\,95.4)};
\node[opbest] at (1.5, -4.8) (h_cg) {CodeGenerate};
\node[opbest] at (1.5, -5.8) (h_te) {Test};
\draw[arr] (h_cg) -- (h_te);

% ==========================================
% HumanEval WITHOUT Contract (right column)
% ==========================================
\node[lbl] at (7.5, -4.0) (h_two) {\textsc{HumanEval} — \textit{w/o $\mathcal{C}$} (Acc\,=\,90.8)};
\node[opworse] at (5.5, -4.8) (h_c0) {CodeGen};
\node[opworse] at (7.5, -4.8) (h_c1) {CodeGen};
\node[opworse] at (9.5, -4.8) (h_c2) {CodeGen};
\node[opworse] at (7.5, -5.8) (h_en) {ScEnsemble};
\node[opworse] at (7.5, -6.8) (h_t2) {Test};
\node[opworse] at (7.5, -7.8) (h_fx) {Custom (fix)};
\draw[arr] (h_c0) -- (h_en);
\draw[arr] (h_c1) -- (h_en);
\draw[arr] (h_c2) -- (h_en);
\draw[arr] (h_en) -- (h_t2);
\draw[arr] (h_t2) -- (h_fx) node[midway, right, font=\tiny] {if fail};

% ==========================================
% Divider lines
% ==========================================
\draw[dashed, gray] (3.5, 0.4) -- (3.5, -3.2);
\draw[dashed, gray] (3.5, -3.6) -- (3.5, -8.2);
\draw[dotted, gray] (-0.2, -3.4) -- (10.8, -3.4);

\end{tikzpicture}
\caption{Synthesized workflow topologies with vs.\ without output contracts. With contracts, the synthesis LLM generates lean, direct pipelines. Without contracts, it compensates for output uncertainty by introducing redundant ensemble voting and error recovery branches---increasing both cost and the number of inter-node interfaces where formatting errors can cascade. Green nodes = with contracts (higher accuracy); red nodes = without contracts (lower accuracy).}
\label{fig:contract_topology}
\end{figure}
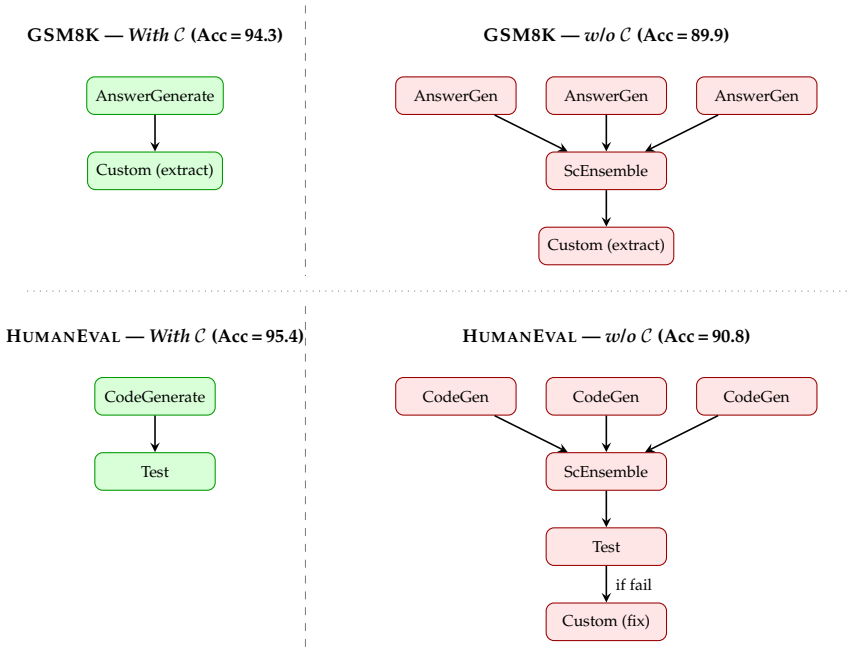

To empirically demonstrate that output contracts influence the \textit{structural design} of synthesized workflows, not merely the terminal output format, we compare the workflow topologies generated with and without contracts ($\mathcal{C}$) across four benchmarks. Figure~\ref{fig:contract_topology} visualizes two representative cases (GSM8K and HumanEval), and Table~\ref{tab:contract_topology} summarizes the structural differences across all four datasets.

\begin{table}[htbp]
\centering
\caption{Structural comparison of synthesized workflows with and without output contracts ($\mathcal{C}$). ``Nodes'' counts total operator invocations; ``Depth'' counts sequential levels. Accuracy from Table~\ref{tab:component_ablation}.}
\label{tab:contract_topology}
\small
\begin{tabular}{l cc cc cc}
\toprule
& \multicolumn{2}{c}{\textbf{With $\mathcal{C}$}} & \multicolumn{2}{c}{\textbf{Without $\mathcal{C}$}} & \multicolumn{2}{c}{\textbf{$\Delta$}} \\
\cmidrule(lr){2-3} \cmidrule(lr){4-5} \cmidrule(lr){6-7}
\textbf{Dataset} & Nodes & Acc & Nodes & Acc & Nodes & Acc \\
\midrule
GSM8K     & 2 & 94.3 & 5 & 89.9 & $+3$  & $-4.4$ \\
MATH      & 7 & 55.1 & 5 & 50.4 & $-2$  & $-4.7$ \\
HumanEval & 2 & 95.4 & 6 & 90.8 & $+4$  & $-4.6$ \\
MBPP      & 3 & 83.3 & 3 & 80.7 & $\pm0$ & $-2.6$ \\
\bottomrule
\end{tabular}
\end{table}

\textbf{Key observations.} (1) On GSM8K and HumanEval, removing contracts causes the synthesizer to triple the node count by adding multi-candidate generation and ensemble voting as hedging mechanisms, yet accuracy \textit{drops} by 4--5 points. (2) On MATH, both versions use ensemble voting, but the contract version selects a more suitable operator (\texttt{Custom} vs.\ \texttt{AnswerGenerate}) and generates more candidates (5 vs.\ 3). (3) MBPP shows near-identical topology in both conditions, suggesting that contracts primarily reshape architectures for tasks with complex output interfaces (math extraction, code validation) rather than tasks where the operator library already enforces output structure (e.g., \texttt{Test} operator in code tasks).

\section{Use of LLM Disclosure}
Large language models were used in this work for (1) evaluation of the proposed methods (as described in the experimental setup) and (2) improving the clarity of the writing.

\end{document}